\documentclass[letterpaper]{article}
\usepackage[letterpaper]{geometry}

\usepackage[utf8]{inputenc}
\usepackage[T1]{fontenc} 
\usepackage[cjk]{kotex}
\usepackage{amsmath}
\usepackage{graphicx}
\usepackage{url}
\usepackage[table]{xcolor}

\usepackage{latexsym}
\usepackage{amssymb} 

\usepackage{hyperref}
\definecolor{darkblue}{rgb}{0, 0, 0.5}
\hypersetup{colorlinks=true,citecolor=darkblue, linkcolor=darkblue, urlcolor=darkblue}

\usepackage{tabularx}
\usepackage{booktabs}

\usepackage{amsthm}
\newtheorem{theorem}{Theorem}
\newtheorem{lemma}{Lemma}
\newtheorem{definition}{Definition}
\newtheorem{proposition}{Proposition}

\usepackage{subcaption}

\usepackage{setspace}
\usepackage{lineno}

\usepackage[authoryear,round]{natbib}
\usepackage{algorithmic,algorithm}
\renewcommand{\algorithmiccomment}[1]{\bgroup\hfill$\triangleright$~#1\egroup}

\usepackage[linguistics]{forest} 
\usepackage{synttree}
\usepackage{subcaption}
\usepackage{tikz-dependency}

\usepackage{CJKutf8}

\usepackage{langsci-gb4e}

\title{\textbf{
\textit{A generalized transition framework for identifying reduction domains in non-binary constituency parsing}
}}
\author{
\begin{tabular}{ccccc}
Jungyeul Park & \qquad & Zihao Huang &\qquad & Eunkyul Leah Jo\\
KAIST && UBC && UBC \\
South Korea && Canada  && Canada \\
{\tt jungyeul@kaist.ac.kr} && \multicolumn{3}{c}{\{\tt hzhhzh1999,eunkyul\}@student.ubc.ca}\\ 
\end{tabular}
}
\date{ 
}

\begin{document}
\maketitle
\begin{abstract}
Direct non-binary constituency parsing must identify an ordered sequence of children at each phrase-level reduction. We extend our earlier delimiter-guided bottom-up construction to a common calculus for top-down, in-order, and bottom-up parsing. At reduction, the rightmost delimiter marks the left boundary, the nonterminal marker supplies the mother label, and the completed items in the delimited segment determine the children and arity. We use a stack invariant to prove soundness and completeness and derive the exact length of the static oracle's action sequence. A projection-and-lifting theorem shows that each delimiter-guided system and its canonical counterpart generate the same set of trees under both the structural and grammar-licensed transition relations. In controlled experiments on PTB and CTB, the delimiter-guided parsers achieve similar labeled accuracy to their canonical counterparts across all traversal orders. For bottom-up parsing, delimiter guidance reduces the number of action classes and increases the number of supervised decisions per tree.
\end{abstract}

\tableofcontents

\doublespacing

\section{Introduction}

{A transition system for constituency parsing must specify how terminals are consumed, how nonterminal labels are introduced, and how partial structures are composed into larger constituents. In the binary shift-reduce tradition, binary reductions combine the two topmost stack items, while separate unary actions construct unary constituents; the action may also specify the mother label and, in lexicalized variants, the head direction. Early classifier-based and deterministic parsers adopted this architecture through actions such as \textsc{Shift}, \textsc{Reduce-Unary}-$X$, and \textsc{Reduce-Left/Right}-$X$ \citep{sagae-lavie-2005-classifier,wang-etal-2006-fast,zhang-clark-2009-transition}. Later neural and globally trained variants retained the same binary operational core while changing the scoring, representation, or search procedure \citep{zhu-etal-2013-fast,watanabe-sumita-2015-transition,liu-zhang-2017-shift}.}

{Binary parsing obtains a compact local reduction operation by transforming a non-binary constituent into a sequence of binary structures. The parser consequently operates over intermediate nodes that must later be removed. Direct non-binary systems instead construct each annotated constituent from its complete child sequence in one phrase-level reduction. Top-down parsing introduces the mother label before its children are processed \citep{dyer-etal-2016-recurrent}; in-order parsing introduces it after the left corner has been completed \citep{liu-zhang-2017-order}; and bottom-up parsing introduces it after all children are available, conventionally selecting them through an action of the form \(\textsc{Reduce-}X\#k\) \citep{gonzalez-rodriguez-2019-faster}.}

{Despite these different traversal orders, every phrase-level reduction constructs the same kind of object: a mother \(X\) over an ordered, nonempty sequence of completed children \(\langle\alpha_1,\ldots,\alpha_k\rangle\). We call this sequence the reduction domain and its cardinality the reduction arity. The systems differ in how they identify that domain. Top-down parsing places a nonterminal marker before the children, in-order parsing places it after the left corner, and arity-specific bottom-up parsing selects a stack suffix through the action parameter \(k\).}

Our earlier bottom-up system showed that a delimiter can identify the child sequence of a reduction and thereby determine its arity \citep{park-etal-2026-non}. We take that construction as the bottom-up case and ask whether the same representation preserves the tree-building capacity of each canonical traversal. The generalized calculus represents a pending phrase-level reduction by a delimiter \(D\), a marker \(\textsc{nt}(X)\), and an ordered sequence of completed children. The traversal parameter \(\tau\in\{\mathrm{TD},\mathrm{IO},\mathrm{BU}\}\) places the marker before all children, after the left corner, or after the complete child sequence. In each case the rightmost delimiter identifies the beginning of the relevant stack segment, the marker supplies the mother label, and the completed items determine the ordered reduction domain.

{We define structural applicability independently of grammar licensing. The structural relation controls the arrangement of terminals, completed items, delimiters, and markers, whereas its grammar-licensed restriction additionally requires every constructed node to instantiate a production of a specified context-free grammar. A stack invariant characterizes the ordering, segmentation, and delimiter discipline of every reachable configuration. Its preservation supports soundness and completeness for all three traversal orders. The static oracle is determined recursively by the gold tree and has \(3n+2m+1\) actions for a tree with \(n\) terminals and \(m\) phrasal nodes.}

{The principal formal result is a global correspondence between each delimiter-guided system and its canonical counterpart. We define projections from complete successful delimiter-guided derivations and operational liftings of complete successful canonical derivations. For top-down parsing, lifting inserts a delimiter with each phrasal marker. For in-order and bottom-up parsing, a delimiter is introduced at a terminal exactly when that terminal is the leftmost terminal of at least one phrasal constituent; reductions along the resulting first-child chain retain the delimiter until its outermost constituent is completed. The projection and lifting preserve terminal yields, mother labels, and ordered child sequences, establishing equality of the generated tree languages under both the structural and grammar-licensed relations.}

{We evaluate the three delimiter-guided instantiations against canonical top-down and in-order parsing and arity-specific bottom-up parsing on the Penn Treebank \citep{marcus-etal-1993-building} and Penn Chinese Treebank \citep{xue-etal-2005-ctb}. Within each traversal order, the systems share the data, preprocessing, neural architecture, optimization procedure, and decoding strategy. The accompanying analyses examine action inventory, constituent arity, and the distinct representational choice made by binary parsing.}

{The resulting account identifies the reduction domain, rather than an arity-labeled action, as the central object of direct non-binary composition. Once a parser configuration identifies a mother label and an ordered child sequence, arity follows from that sequence and need not be represented as an independent transition parameter.}

\section{Background and notation}
\label{sec:background}

Let \(w=w_1\cdots w_n\) be an input sentence. A constituency tree over
\(w\) is a rooted ordered tree whose leaves are the terminals
\(w_1,\ldots,w_n\). Each terminal is immediately dominated by a
preterminal labeled with a part-of-speech category, and each higher
internal node is labeled with a phrasal nonterminal. We assume
projectivity, so the yield of every node is contiguous.

Let \(\mathcal{N}\) and \(\mathcal{P}\) be the finite sets of phrasal and
preterminal labels, respectively, with
\(\mathcal{N}\cap\mathcal{P}=\varnothing\). A completed preterminal item
has the form \(P(w_i)\), where \(P\in\mathcal{P}\). A completed phrasal
item has the form
\[
X(\alpha_1,\ldots,\alpha_k),
\]
where \(X\in\mathcal{N}\), \(k\geq1\), and each \(\alpha_j\) is a
completed preterminal or phrasal item. The sequence
\(\alpha_1,\ldots,\alpha_k\) is contiguous when its members cover
adjacent, nonoverlapping yields in left-to-right order.

In the stack schemas below, \(\sigma\) denotes an arbitrary stack prefix
and \(\mid\) denotes stack concatenation. The transition
\(\textsc{Nt}(X)\) introduces a marker \(\textsc{nt}(X)\). We use
\(\textsc{Nt}(X)\) for the transition and \(\textsc{nt}(X)\) for the
stack symbol that it introduces.

\begin{definition}[Reduction domain and arity]
For a phrase-level construction
\(X(\alpha_1,\ldots,\alpha_k)\), the ordered nonempty sequence
\[
\langle\alpha_1,\ldots,\alpha_k\rangle
\]
is its \emph{reduction domain}. Its left boundary is the left edge of
the yield of \(\alpha_1\), and its arity is \(k\).
\end{definition}

\subsection{Canonical direct non-binary systems}

Top-down, in-order, and bottom-up non-binary parsing construct the same
phrase-level object but encode its reduction domain differently. Their
characteristic reducible stack suffixes and reduction effects are
\[
\begin{array}{rcll}
\mathrm{TD}:&
\sigma\mid\textsc{nt}(X)\mid
\alpha_1\mid\cdots\mid\alpha_k
&\xrightarrow{\textsc{Reduce}}&
\sigma\mid X(\alpha_1,\ldots,\alpha_k),
\\[1mm]
\mathrm{IO}:&
\sigma\mid\alpha_1\mid\textsc{nt}(X)\mid
\alpha_2\mid\cdots\mid\alpha_k
&\xrightarrow{\textsc{Reduce}}&
\sigma\mid X(\alpha_1,\ldots,\alpha_k),
\\[1mm]
\mathrm{BU}:&
\sigma\mid\alpha_1\mid\cdots\mid\alpha_k
&\xrightarrow{\textsc{Reduce-}X\#k}&
\sigma\mid X(\alpha_1,\ldots,\alpha_k),
\end{array}
\]
where \(X\in\mathcal{N}\) and \(k\geq1\). For \(k=1\), the sequence
\(\alpha_2,\ldots,\alpha_k\) in the in-order schema is empty. {These schemas display only the stack effects relevant to phrase-level reduction; the complete canonical configurations, transitions, and applicability conditions are specified in Appendix~\ref{app:canonical-transition-systems}.}

In top-down parsing, the mother label is introduced before any child is
constructed \citep{dyer-etal-2016-recurrent}. The unmatched marker
supplies \(X\) and identifies the beginning of the reduction domain.
The arity is therefore the number of completed children above that
marker.

In in-order parsing, the marker is introduced after the completed left
corner \(\alpha_1\) and before the remaining children
\citep{liu-zhang-2017-order}. The reduction domain consists of the left
corner immediately below the marker and the completed children above
it. Top-down and in-order parsing thus recover arity from their marker
configurations rather than from an arity-specific action.

In arity-specific bottom-up parsing, all children are completed before
the mother label is introduced
\citep{gonzalez-rodriguez-2019-faster}. Because no marker identifies
the left boundary of the child sequence, the parameter \(k\) in
\(\textsc{Reduce-}X\#k\) selects the top \(k\) completed stack items.
The action therefore jointly specifies the mother label and reduction
arity.

\subsection{Structural and grammar-licensed transitions}
\label{sec:background-grammar}

Let \(G\) denote the context-free grammar induced by the treebank, with
production set \(R\) and start symbol \(S\in\mathcal{N}\). A completed
preterminal \(P(w_i)\) instantiates a production
\(P\rightarrow w_i\). A completed phrasal item
\(X(\alpha_1,\ldots,\alpha_k)\) instantiates
\[
X\rightarrow Y_1\cdots Y_k,
\]
where \(Y_j\) is the label of \(\alpha_j\).

For each traversal order
\(\tau\in\{\mathrm{TD},\mathrm{IO},\mathrm{BU}\}\), the structural
relation \(\Rightarrow_{\tau}\) enforces the operational
conditions of the transition system: label typing, input and completion
conditions, traversal-specific stack patterns, delimiter and marker
placement, and the contiguity and ordering of completed children. It
does not require a constructed item to instantiate a production in
\(R\).

The grammar-licensed relation \(\Rightarrow_{\tau,G}\) restricts \(\Rightarrow_{\tau}\) by requiring every constructed item to be licensed by \(R\). Formally, for any configurations \(c\) and \(c'\),
\[
c\Rightarrow_{\tau,G}c'
\quad\Longrightarrow\quad
c\Rightarrow_{\tau}c',
\]
or equivalently,
\[
\Rightarrow_{\tau,G}\subseteq\Rightarrow_{\tau}.
\]
A preterminal construction requires \(P\rightarrow w_i\in R\), and a phrase-level construction requires \(X\rightarrow Y_1\cdots Y_k\in R\). Nonconstructive transitions coincide under the two relations. This distinction separates structural well-formedness and projectivity from membership in the treebank-induced grammar. The neural decoders select actions using implementation-specific legality checks without consulting \(R\). The formal results concern \(\Rightarrow_{\tau}\) and \(\Rightarrow_{\tau,G}\). Appendix~\ref{app:experimental-details} specifies where the implemented checks differ from those relations.

\subsection{Relation to our earlier bottom-up system}
\label{sec:relation-prior}

Our earlier work introduced delimiter-guided non-binary bottom-up
parsing \citep{park-etal-2026-non}. Its phrase-level reducible
suffix has the form
\[
D\mid\alpha_1\mid\cdots\mid\alpha_k\mid\textsc{nt}(X).
\]
The delimiter \(D\) identifies the left boundary, the marker supplies
the mother label, and the completed items between them determine the
ordered reduction domain and its arity. The actions \textsc{d-Shift}
and \textsc{d-Reduce} respectively introduce this boundary and preserve
it across constituents sharing the same left edge. The earlier experiments established the accompanying trade-off between fewer action classes and longer bottom-up derivations.

\citet{yang-tu-2022-bottom} also identify constituent boundaries without predicting a child count. Their bottom-up parser traverses spans in postorder and predicts the other endpoint of the next span from a cursor at a boundary shared with the preceding span. In the delimiter calculus, bottom-up parsing establishes a boundary when shifting its first word and can retain that boundary across multiple reductions. The reduction reads the completed children from the resulting stack segment. The separation of boundary placement, mother-label introduction, and reduction lets us compare their timing across traversal orders.

The present calculus retains this separation and changes when the boundary and label are introduced. The formal question is whether those choices admit a constructive correspondence with the canonical derivations of each traversal order.

Canonical top-down transition systems do not require an independent delimiter because an unmatched nonterminal marker already identifies the left boundary of the reduction domain. In the generalized calculus, a phrasal \(\textsc{Nt}(X)\) explicitly separates these roles by jointly introducing \(D\mid\textsc{nt}(X)\): \(D\) records the boundary, while \(\textsc{nt}(X)\) records the mother label. The delimiter is not introduced by a separate predicted action and is consumed together with its marker by the corresponding \textsc{Reduce}; erasing \(D\) recovers the canonical top-down stack schema. By contrast, in-order and bottom-up parsing introduce the mother marker only after part or all of the child sequence has been constructed, so \textsc{d-Shift} must establish the boundary earlier and \textsc{d-Reduce} may preserve it across constituents sharing that boundary.

\section{Generalized delimiter-guided transition operators}
\label{sec:generalized-transition-operators}

{For each traversal order \(\tau\in\{\mathrm{TD},\mathrm{IO},\mathrm{BU}\}\), a parser configuration is \(c=\langle\sigma,i,f\rangle\), where \(\sigma\) is a stack, \(i\in\{1,\ldots,n+1\}\) is the index of the next input token, and \(f\in\{0,1\}\) is the completion flag. The initial configuration is \(\langle\epsilon,1,0\rangle\). A successful final configuration has the form \(\langle T,n+1,1\rangle\), where \(T\) is a completed tree rooted in the start symbol \(S\) with yield \(w_1\cdots w_n\).}

{The stack may contain delimiters \(D\), markers \(\textsc{nt}(L)\) with \(L\in\mathcal{N}\cup\mathcal{P}\), bare terminals, and completed tree items. Each action below is a partial function on configurations, and the union of these functions defines the structural relation \(\Rightarrow_{\tau}\). The grammar-licensed relation \(\Rightarrow_{\tau,G}\) additionally imposes the production conditions stated below. The top-down system excludes \textsc{d-Shift} and \textsc{d-Reduce}; the in-order and bottom-up systems admit both. Every action requires \(f=0\), and no action is applicable after \(f=1\).}

{In every transition schema, \(\sigma\) denotes an arbitrary stack prefix and the displayed elements constitute the complete suffix above it. In an in-order or bottom-up phrasal \textsc{Nt} schema, the displayed \(D\) is the rightmost active delimiter and the suffix beginning at \(D\) contains no other delimiter or marker. In a phrase-level reduction schema, the displayed \(D\) is likewise the rightmost active delimiter, and its suffix contains no other delimiter and no marker other than the one displayed. The items \(\alpha_1,\ldots,\alpha_k\) are completed tree items with nonempty consecutive yields in left-to-right order. The delimiter introduced by a top-down phrasal \textsc{Nt} action is fresh.}

{An unfinished preterminal construction must be completed before any other construction begins. In the top-down order, a stack-final \(\textsc{nt}(P)\) admits only the corresponding \textsc{Shift}, and the resulting suffix \(\textsc{nt}(P)\mid w_{i-1}\) admits only \textsc{u-Reduce}. In the in-order and bottom-up orders, a stack-final bare terminal \(w_{i-1}\) admits only an instance of \(\textsc{Nt}(P)\), and the resulting suffix \(w_{i-1}\mid\textsc{nt}(P)\) admits only \textsc{u-Reduce}. Thus no transition places a delimiter, marker, terminal, or completed item above an unfinished preterminal construction.}

\subsection{Operator inventory}
\label{operator-inventory}

\paragraph{\textsc{Shift}}

{For \(i\leq n\), \textsc{Shift} consumes \(w_i\) without introducing a delimiter:}
\[
{
\begin{array}{llll}
\mathrm{TD}:&
\langle\sigma\mid\textsc{nt}(P),i,0\rangle
&\Rightarrow_{\mathrm{TD}}&
\langle\sigma\mid\textsc{nt}(P)\mid w_i,i+1,0\rangle,
\\[1mm]
\mathrm{IO/BU}:&
\langle\sigma,i,0\rangle
&\Rightarrow_{\tau}&
\langle\sigma\mid w_i,i+1,0\rangle,
\qquad \tau\in\{\mathrm{IO},\mathrm{BU}\}.
\end{array}
}
\]
{In the top-down schema, \(P\in\mathcal{P}\) is the pending preterminal label.}

\paragraph{\textsc{d-Shift}}

For the in-order and bottom-up systems, \textsc{d-Shift} is structurally applicable when the derivation is unfinished (\(f=0\)), a word remains in the input (\(i\leq n\)), and no preterminal construction is pending. It pushes a fresh delimiter followed by the bare terminal \(w_i\) and advances the input by one position.
\[
{
\langle\sigma,i,0\rangle
\Rightarrow_{\tau}
\langle\sigma\mid D\mid w_i,i+1,0\rangle.
}
\]
The ensuing \textsc{Nt}(\(P\)) and \textsc{u-Reduce} leave the delimiter immediately below the completed preterminal. On the gold derivation, the oracle chooses \textsc{d-Shift} exactly when \(w_i\) is the leftmost terminal of at least one phrasal constituent, and chooses plain \textsc{Shift} elsewhere. A single delimiter serves all phrases that begin at that word, even when an earlier delimiter is still active. Top-down parsing introduces each delimiter with a phrasal \textsc{Nt} action before building its children.

\paragraph{\textsc{Nt}(\(L\))}

{For \(L\in\mathcal{N}\cup\mathcal{P}\), \textsc{Nt}(\(L\)) introduces the marker \(\textsc{nt}(L)\). The expressions \textsc{Nt}(\(P\)) and \textsc{Nt}(\(X\)) below are typed schema instances of this single label-parameterized family, with \(P\in\mathcal{P}\) and \(X\in\mathcal{N}\). Each concrete label \(L\) nevertheless instantiates a distinct predicted action type.}

{For a preterminal label \(P\in\mathcal{P}\), the marker position depends on the traversal order:}
\[
{
\begin{array}{llll}
\mathrm{TD}:&
\langle\sigma,i,0\rangle
&\Rightarrow_{\mathrm{TD}}&
\langle\sigma\mid\textsc{nt}(P),i,0\rangle,
\qquad i\leq n,
\\[1mm]
\mathrm{IO/BU}:&
\langle\sigma\mid w_{i-1},i,0\rangle
&\Rightarrow_{\tau}&
\langle\sigma\mid w_{i-1}\mid\textsc{nt}(P),i,0\rangle,
\\
&&&
\tau\in\{\mathrm{IO},\mathrm{BU}\},\qquad 2\leq i\leq n+1.
\end{array}
}
\]
{Thus, top-down parsing introduces the preterminal label before consuming its terminal, whereas in-order and bottom-up parsing introduce it afterward. Preterminal marker introduction does not introduce a delimiter.}

{For a phrasal label \(X\in\mathcal{N}\), the traversal order determines how the delimiter is obtained and where the marker is placed:}
\[
{
\begin{array}{llll}
\mathrm{TD}:&
\langle\sigma,i,0\rangle
&\Rightarrow_{\mathrm{TD}}&
\langle\sigma\mid D\mid\textsc{nt}(X),i,0\rangle,
\qquad i\leq n,
\\[1mm]
\mathrm{IO}:&
\langle\sigma\mid D\mid\alpha_1,i,0\rangle
&\Rightarrow_{\mathrm{IO}}&
\langle\sigma\mid D\mid\alpha_1\mid\textsc{nt}(X),i,0\rangle,
\\[1mm]
\mathrm{BU}:&
\langle\sigma\mid D\mid\alpha_1\mid\cdots\mid\alpha_k,i,0\rangle
&\Rightarrow_{\mathrm{BU}}&
\langle\sigma\mid D\mid\alpha_1\mid\cdots\mid\alpha_k\mid\textsc{nt}(X),i,0\rangle,
\qquad k\geq1.
\end{array}
}
\]
{In the top-down system, \textsc{Nt}(\(X\)) jointly introduces a fresh delimiter and the marker \(\textsc{nt}(X)\) before any child is constructed. The corresponding \textsc{Reduce} removes both symbols, so neither \textsc{d-Shift} nor \textsc{d-Reduce} is required. In the in-order and bottom-up systems, the active delimiter has already been introduced by \textsc{d-Shift}, possibly retained by \textsc{d-Reduce}, and \textsc{Nt}(\(X\)) introduces only the mother-label marker after the left corner or the complete child sequence, respectively.}

\paragraph{\textsc{u-Reduce}}

{The action \textsc{u-Reduce} completes a preterminal:}
\[
{
\begin{array}{llll}
\mathrm{TD}:&
\langle\sigma\mid\textsc{nt}(P)\mid w_{i-1},i,0\rangle
&\Rightarrow_{\mathrm{TD}}&
\langle\sigma\mid P(w_{i-1}),i,0\rangle,
\\[1mm]
\mathrm{IO/BU}:&
\langle\sigma\mid w_{i-1}\mid\textsc{nt}(P),i,0\rangle
&\Rightarrow_{\tau}&
\langle\sigma\mid P(w_{i-1}),i,0\rangle,
\\
&&&
\tau\in\{\mathrm{IO},\mathrm{BU}\},\qquad 2\leq i\leq n+1.
\end{array}
}
\]
{Under \(\Rightarrow_{\tau,G}\), the construction additionally requires \(P\rightarrow w_{i-1}\in R\). The action applies only to \(P\in\mathcal{P}\); a unary phrasal constituent is constructed by a phrase-level reduction with one completed child.}

\paragraph{\textsc{Reduce}}

{Let \(X\in\mathcal{N}\), let \(k\geq1\), and let \(Y_j\) be the root label of \(\alpha_j\). The traversal-specific reductions are}
\[
{
\begin{array}{llll}
\mathrm{TD}:&
\langle\sigma\mid D\mid\textsc{nt}(X)\mid\alpha_1\mid\cdots\mid\alpha_k,i,0\rangle
&\Rightarrow_{\mathrm{TD}}&
\langle\sigma\mid X(\alpha_1,\ldots,\alpha_k),i,0\rangle,
\\[1mm]
\mathrm{IO}:&
\langle\sigma\mid D\mid\alpha_1\mid\textsc{nt}(X)\mid\alpha_2\mid\cdots\mid\alpha_k,i,0\rangle
&\Rightarrow_{\mathrm{IO}}&
\langle\sigma\mid X(\alpha_1,\ldots,\alpha_k),i,0\rangle,
\\[1mm]
\mathrm{BU}:&
\langle\sigma\mid D\mid\alpha_1\mid\cdots\mid\alpha_k\mid\textsc{nt}(X),i,0\rangle
&\Rightarrow_{\mathrm{BU}}&
\langle\sigma\mid X(\alpha_1,\ldots,\alpha_k),i,0\rangle.
\end{array}
}
\]
{For \(k=1\), the sequence following the marker in the in-order schema is empty. Each reduction removes the displayed delimiter and marker and replaces the ordered reduction domain with \(X(\alpha_1,\ldots,\alpha_k)\). Under \(\Rightarrow_{\tau,G}\), it additionally requires}
\[
{X\rightarrow Y_1\cdots Y_k\in R.}
\]

\paragraph{\textsc{d-Reduce}}

For the in-order and bottom-up systems, \textsc{d-Reduce} has the same structural preconditions as phrase-level \textsc{Reduce}. The suffix above the rightmost active delimiter must consist of one phrasal marker and one or more completed children whose yields are consecutive from left to right. The in-order marker follows exactly one child, with the remaining children above it. The bottom-up marker follows all children and is at the top of the stack. The action removes the marker, replaces the children with the new phrase, and leaves the same delimiter immediately below it. The input position stays fixed.
\[
{
\begin{array}{llll}
\mathrm{IO}:&
\langle\sigma\mid D\mid\alpha_1\mid\textsc{nt}(X)\mid\alpha_2\mid\cdots\mid\alpha_k,i,0\rangle
&\Rightarrow_{\mathrm{IO}}&
\langle\sigma\mid D\mid X(\alpha_1,\ldots,\alpha_k),i,0\rangle,
\\[1mm]
\mathrm{BU}:&
\langle\sigma\mid D\mid\alpha_1\mid\cdots\mid\alpha_k\mid\textsc{nt}(X),i,0\rangle
&\Rightarrow_{\mathrm{BU}}&
\langle\sigma\mid D\mid X(\alpha_1,\ldots,\alpha_k),i,0\rangle,
\qquad k\geq1.
\end{array}
}
\]
The two reductions construct the same phrase and differ only in whether the delimiter remains available for a later reduction. On the gold derivation, the oracle chooses \textsc{d-Reduce} when a higher phrase shares the same left boundary. This occurs exactly when the new phrase is the first child of its parent. The oracle chooses \textsc{Reduce} at the outermost phrase beginning at that boundary. Under \(\Rightarrow_{\tau,G}\), both actions require \(X\rightarrow Y_1\cdots Y_k\in R\).

{Whenever \textsc{Reduce} or \textsc{d-Reduce} is structurally applicable, the rightmost active delimiter fixes the beginning of the reduction suffix, its single marker determines the mother label, and all completed items in that suffix determine the ordered child sequence and its arity. Grammar licensing may reject the resulting production, but it does not alter the reduction domain.}

\paragraph{\textsc{Finish}}

{After all input has been consumed, \textsc{Finish} applies exactly when the stack contains one completed tree rooted in \(S\):}
\[
{
\langle S(\alpha_1,\ldots,\alpha_k),n+1,0\rangle
\Rightarrow_{\tau}
\langle S(\alpha_1,\ldots,\alpha_k),n+1,1\rangle,
\qquad k\geq1.
}
\]
{The same condition holds under \(\Rightarrow_{\tau,G}\). The target configuration is final and admits no further transition.}

\subsection{Traversal order and static-oracle choices}

{For a constituent with \(k\geq1\) children, define}
\[
{
\operatorname{Order}_{\tau}(k)=
\begin{cases}
0, & \tau=\mathrm{TD},\\
1, & \tau=\mathrm{IO},\\
k, & \tau=\mathrm{BU}.
\end{cases}
}
\]
{This value is the number of completed children preceding the mother-label marker. Let \(r=\operatorname{Order}_{\tau}(k)\). The complete reducible suffix for \(X(\alpha_1,\ldots,\alpha_k)\) has the uniform form}
\[
{
D\mid
\alpha_1\mid\cdots\mid\alpha_r\mid
\textsc{nt}(X)\mid
\alpha_{r+1}\mid\cdots\mid\alpha_k,
}
\]
{where an empty child subsequence is omitted. Thus, the marker precedes all children in the top-down order, follows the left corner in the in-order order, and follows all children in the bottom-up order. The delimiter precedes the complete reduction domain in every order.}

{For the in-order and bottom-up static oracles, let \(p_i\) be the preterminal dominating \(w_i\), and let \(b(w_i)\) be the highest phrasal ancestor of \(p_i\) reached by following first-child edges only. The value \(b(w_i)\) is defined exactly when \(w_i\) is the leftmost terminal of at least one phrasal constituent. Define the terminal-consuming action}
\[
{
A_i=
\begin{cases}
\textsc{d-Shift}, & b(w_i)\text{ is defined},\\
\textsc{Shift}, & b(w_i)\text{ is undefined}.
\end{cases}
}
\]
{When \(A_i=\textsc{d-Shift}\), the introduced delimiter marks the common left boundary of the phrasal constituents on the first-child path from \(p_i\) to \(b(w_i)\). The boundary target \(b(w_i)\) is used only to define the static oracle and is not stored in the parser configuration.}
The in-order and bottom-up oracle code also checks whether \(w_i\) is sentence-initial or its preterminal and the preceding word's preterminal have different parents. This check follows from the existence of \(b(w_i)\) on a well-typed tree. If \(w_i\) is later in the sentence and \(b(w_i)\) exists, \(p_i\) is the first child of its parent and the preceding word lies outside that parent's yield.

{For a phrasal node occurrence \(v=X(T_1,\ldots,T_k)\), let \(w_{\ell(v)}\) be its leftmost terminal. Its static-oracle reduction action is}
\[
{
R_v=
\begin{cases}
\textsc{Reduce},
& v=b(w_{\ell(v)}),\\
\textsc{d-Reduce},
& v\text{ is a proper descendant of }b(w_{\ell(v)})
\text{ on the same first-child path}.
\end{cases}
}
\]
{These cases are exhaustive because \(v\) lies on the first-child path from the preterminal over \(w_{\ell(v)}\) to \(b(w_{\ell(v)})\). Reductions below the boundary target retain the delimiter, whereas reduction of the boundary target consumes it. The in-order and bottom-up static oracles use the same \(A_i\), \(b(w_i)\), and node-indexed reduction actions \(R_v\); they differ only in whether \(\textsc{nt}(X)\) is introduced after the left corner or after the complete child sequence.} At test time the decoder predicts the shift and reduction variants from the current configuration under its own legality checks (Appendix~\ref{app:experimental-details}).

In Figure~\ref{fig:ms-haag}, \textsc{d-Shift} on \textit{Ms.} establishes the shared boundary of \(NP_1\), \(S\), and \(TOP\). Plain \textsc{Shift} on \textit{Haag} leaves that boundary in place. Applying ordinary \textsc{Reduce} at \(NP_1\) would discard the boundary still needed by \(S\) and \(TOP\). The oracle therefore uses \textsc{d-Reduce} at \(NP_1\) and \(S\), then consumes the delimiter with \textsc{Reduce} at \(TOP\).

\subsection{Reachability and stack invariant}

{A configuration is structurally reachable if it is the initial configuration or follows from a structurally reachable configuration by one step of \(\Rightarrow_{\tau}\). Unless stated otherwise, \emph{reachable} means structurally reachable. Because \(\Rightarrow_{\tau,G}\subseteq\Rightarrow_{\tau}\), every configuration reachable under the grammar-licensed relation is also structurally reachable.}

{A delimiter is \emph{active} while it remains on the stack. Each active delimiter determines a \emph{delimiter segment}: the portion of the stack above that delimiter and below the next active delimiter, or below the top of the stack if no delimiter occurs above it. Its boundary is the input boundary at which it was introduced; this analytical attribute is not part of the parser configuration or action alphabet. The stack invariant consists of the following conditions.}

\begin{enumerate}

\item {After delimiters and markers are ignored, the yields of the completed tree items, together with an optional bare terminal, form a disjoint left-to-right partition of the consumed input prefix \(w_1\cdots w_{i-1}\). Every completed tree item has a nonempty contiguous yield.}

\item {At most one preterminal construction is unfinished. In a configuration whose next-input index is \(i\), its stack-final form is \(\textsc{nt}(P)\), or \(\textsc{nt}(P)\mid w_{i-1}\), for \(\tau=\mathrm{TD}\), and \(w_{i-1}\), or \(w_{i-1}\mid\textsc{nt}(P)\), for \(\tau\in\{\mathrm{IO},\mathrm{BU}\}\), with the first top-down form occurring before the corresponding terminal is consumed. No other bare terminal or preterminal marker occurs on the stack.}

\item {Within every delimiter segment, completed tree items occur in left-to-right order and have consecutive yields. The segment contains at most one phrasal marker. In the top-down order, this marker immediately follows the delimiter and precedes every completed item in the segment. In the in-order order, exactly one completed item precedes the marker. In the bottom-up order, a nonempty sequence of completed items precedes the marker; completed items may also follow it.}

\item {The boundaries of active delimiters are nondecreasing in bottom-to-top stack order. Every completed item below an active delimiter has yield strictly to the left of its boundary. If a delimiter segment contains a yield-bearing symbol, its leftmost such symbol begins at the delimiter's boundary.}

\end{enumerate}

\section{Formal properties}
\label{sec:formal-properties}

\begin{lemma}[Invariant preservation]
\label{lem:invariant-preservation}
{For every \(\tau\in\{\mathrm{TD},\mathrm{IO},\mathrm{BU}\}\), every configuration reachable under \(\Rightarrow_{\tau}\) satisfies the stack invariant. Consequently, every configuration reachable under \(\Rightarrow_{\tau,G}\) also satisfies it.}
\end{lemma}

\begin{proof}
{The proof is by induction on derivation length. The initial configuration \(\langle\epsilon,1,0\rangle\) satisfies all four conditions. Suppose that a reachable configuration \(c\) satisfies the invariant and that an applicable action maps \(c\) to \(c'\).}

{A \textsc{Shift} consumes the next terminal and places it in the traversal-specific unfinished preterminal configuration. A \textsc{d-Shift} has the same terminal effect and introduces a delimiter immediately before the terminal, above every active delimiter. Neither action changes the yields or relative order of existing completed items. The applicability conditions for unfinished preterminal constructions ensure that no unrelated action can intervene before the bare terminal is completed.}

{A preterminal \textsc{Nt}(\(P\)) places its marker before the terminal in the top-down order and after the terminal in the in-order and bottom-up orders. The corresponding \textsc{u-Reduce} replaces the terminal and marker with \(P(w_{i-1})\), whose yield is the singleton \(w_{i-1}\). If a delimiter immediately precedes the terminal, both actions leave it in place, and it remains anchored immediately before the resulting preterminal. Thus the consumed-prefix partition and delimiter boundaries are preserved and the unique unfinished preterminal construction is removed.}

{A top-down phrasal \textsc{Nt}(\(X\)) introduces \(D\mid\textsc{nt}(X)\) at the current input boundary, opening a segment whose marker precedes every completed item subsequently constructed in that segment. A \textsc{d-Shift} opens an in-order or bottom-up segment at the same boundary as its terminal. Both actions place the new delimiter above every active delimiter, and their boundary cannot precede the boundary of any delimiter below it. An in-order phrasal marker follows the single completed left corner of its segment, whereas a bottom-up phrasal marker follows a nonempty sequence of completed items. A phrase-level reduction replaces the marker and its ordered completed children with their mother item. Their consecutive yields have a contiguous union. \textsc{Reduce} removes the delimiting \(D\); \textsc{d-Reduce} retains it immediately before a completed item with the same left boundary. These effects preserve the ordering and anchoring of all remaining delimiters.}

{\textsc{Finish} changes only the completion flag. Every applicable transition therefore preserves the invariant. The grammar-licensed claim follows from \(\Rightarrow_{\tau,G}\subseteq\Rightarrow_{\tau}\).}
\end{proof}

\begin{theorem}[Soundness]
\label{thm:soundness}
{For every traversal order \(\tau\), every successful derivation under \(\Rightarrow_{\tau}\) yields a well-typed projective constituency tree rooted in \(S\) whose yield is \(w_1\cdots w_n\). If the derivation is under \(\Rightarrow_{\tau,G}\), the resulting tree is licensed by \(G\).}
\end{theorem}

\begin{proof}
{Induct on construction actions. A \textsc{u-Reduce} constructs \(P(w_{i-1})\), where \(P\in\mathcal{P}\), with singleton yield \(w_{i-1}\). If \(\alpha_1,\ldots,\alpha_k\) are well-typed projective trees with consecutive yields, a phrase-level reduction with \(X\in\mathcal{N}\) constructs \(X(\alpha_1,\ldots,\alpha_k)\). Lemma~\ref{lem:invariant-preservation} ensures that the children occur in left-to-right order over consecutive yields, so their mother is also well typed and projective.}

{Terminal-consuming actions read each token exactly once and in input order, and no construction action deletes, duplicates, or reorders a terminal yield. A successful \textsc{Finish} requires complete input consumption and a stack containing exactly one completed tree rooted in \(S\); its yield is therefore \(w_1\cdots w_n\). Under \(\Rightarrow_{\tau,G}\), each preterminal construction requires \(P\rightarrow w_{i-1}\in R\), and each phrase-level construction requires \(X\rightarrow Y_1\cdots Y_k\in R\), where \(Y_j\) is the root label of the \(j\)-th child. Induction over construction actions consequently licenses every node of the final tree.}
\end{proof}

\begin{definition}[Static oracle]
\label{def:static-oracle}
{Fix a projective gold tree \(T\). Using the terminal actions \(A_i\) and the node-indexed phrasal reduction actions \(R_v\) defined in Section~\ref{sec:generalized-transition-operators}, define the node-level oracle sequences recursively by}
\[
{
\begin{aligned}
o_{\mathrm{TD}}(P(w_i))
&=
\textsc{Nt}(P)\cdot
\textsc{Shift}\cdot
\textsc{u-Reduce},
\\
o_{\mathrm{TD}}(X(T_1,\ldots,T_k))
&=
\textsc{Nt}(X)\cdot
o_{\mathrm{TD}}(T_1)\cdots o_{\mathrm{TD}}(T_k)\cdot
\textsc{Reduce},
\\[1mm]
o_{\mathrm{IO}}(P(w_i))
&=
A_i\cdot
\textsc{Nt}(P)\cdot
\textsc{u-Reduce},
\\
o_{\mathrm{IO}}(X(T_1,\ldots,T_k))
&=
o_{\mathrm{IO}}(T_1)\cdot
\textsc{Nt}(X)\cdot
o_{\mathrm{IO}}(T_2)\cdots o_{\mathrm{IO}}(T_k)\cdot
R_{X(T_1,\ldots,T_k)},
\\[1mm]
o_{\mathrm{BU}}(P(w_i))
&=
A_i\cdot
\textsc{Nt}(P)\cdot
\textsc{u-Reduce},
\\
o_{\mathrm{BU}}(X(T_1,\ldots,T_k))
&=
o_{\mathrm{BU}}(T_1)\cdots o_{\mathrm{BU}}(T_k)\cdot
\textsc{Nt}(X)\cdot
R_{X(T_1,\ldots,T_k)}.
\end{aligned}
}
\]
{Here \(\cdot\) denotes sequence concatenation, and the sequence \(o_{\mathrm{IO}}(T_2)\cdots o_{\mathrm{IO}}(T_k)\) is empty when \(k=1\). The complete oracle sequence is}
\[
{
O_{\tau}(T)=o_{\tau}(T)\cdot\textsc{Finish}.
}
\]
\end{definition}

{For a fixed projective tree \(T\), its ordered node structure determines every \(A_i\) and every node-indexed reduction action \(R_v\), so these recursive clauses uniquely determine \(O_{\tau}(T)\).}

\begin{lemma}[Boundary lifecycle]
\label{lem:boundary-lifecycle}
{Let \(\tau\in\{\mathrm{IO},\mathrm{BU}\}\), and construct a projective tree \(T\) in recursive in-order when \(\tau=\mathrm{IO}\) or recursive postorder when \(\tau=\mathrm{BU}\). Let \(w_i\) be the leftmost terminal of at least one phrasal constituent. The phrasal constituents with left boundary \(i\) form a chain \(T_1,\ldots,T_q\), ordered from innermost to outermost. If \(w_i\) is consumed by \textsc{d-Shift}, the resulting delimiter is the rightmost active delimiter when each \(T_j\) is reduced. It is retained by \textsc{d-Reduce} for \(j<q\) and consumed by \textsc{Reduce} for \(j=q\); every delimiter introduced strictly inside the yield of \(T_j\) is consumed before \(T_j\) is reduced.}
\end{lemma}

\begin{proof}
{Constituent yields are contiguous, so any two phrasal constituents with the same left boundary are nested and therefore form a first-child chain. The \textsc{d-Shift} consuming \(w_i\) places a delimiter immediately before that terminal, and completion of its preterminal leaves the delimiter immediately before the resulting item.}

{Prove the claim simultaneously for all boundaries by induction on the yield length of the outermost member of the chain. If that yield contains one terminal, no later child can introduce a delimiter; the traversal-specific schemas successively reduce the unary chain and establish the claim. Otherwise, a child not lying on the first-child chain has a strictly later left boundary and a strictly shorter yield. By the induction hypothesis, every delimiter opened within that child is removed by the outermost constituent sharing its boundary before the traversal returns from the child. Consequently, when \(T_j\) is ready for reduction, no delimiter introduced inside a later child remains above the delimiter at \(i\). The latter is therefore the rightmost active delimiter, and its segment contains exactly the completed children of \(T_j\), with the marker in the position prescribed by the traversal order. Thus \textsc{d-Reduce} is applicable for \(j<q\) and retains the delimiter immediately before \(T_j\). The same argument then applies to \(T_{j+1}\). At \(T_q\), \textsc{Reduce} is applicable to the same delimited domain and removes the delimiter.}
\end{proof}

\begin{theorem}[Completeness]
\label{thm:completeness}
{Let \(T\) be a well-typed projective constituency tree over \(w_1\cdots w_n\), with preterminal labels in \(\mathcal{P}\), phrasal labels in \(\mathcal{N}\), and root label \(S\). For every traversal order \(\tau\), \(O_{\tau}(T)\) is a successful derivation under \(\Rightarrow_{\tau}\). If \(T\) is licensed by \(G\), the same sequence is successful under \(\Rightarrow_{\tau,G}\).}
\end{theorem}

\begin{proof}
{We prove by structural induction that \(o_{\tau}(U)\) constructs every subtree \(U\) over its complete yield. For \(\tau\in\{\mathrm{IO},\mathrm{BU}\}\), the induction claim additionally records the delimiter at the left boundary of \(U\): every delimiter introduced strictly inside the yield of \(U\) has been consumed when \(U\) is completed, while the left-boundary delimiter remains immediately below \(U\) precisely when \(U\) is a proper descendant of its boundary target. It is consumed when \(U\) is that target.}

{For a preterminal \(P(w_i)\), the top-down sequence}
\[
{
\textsc{Nt}(P)\cdot\textsc{Shift}\cdot\textsc{u-Reduce}
}
\]
{constructs \(P(w_i)\). In the in-order and bottom-up orders,}
\[
{
A_i\cdot\textsc{Nt}(P)\cdot\textsc{u-Reduce}
}
\]
{constructs the same item. If \(A_i=\textsc{Shift}\), no delimiter is introduced. If \(A_i=\textsc{d-Shift}\), the result is \(D\mid P(w_i)\), and \(D\) is retained for the phrasal first-child chain beginning at \(w_i\). This establishes the strengthened base case.}

{Let \(U=X(U_1,\ldots,U_k)\), and assume the claim for its children. In the top-down order, \textsc{Nt}(\(X\)) introduces \(D\mid\textsc{nt}(X)\) before their oracle sequences. Once all children have been completed, the stack has suffix}
\[
{
D\mid\textsc{nt}(X)\mid U_1\mid\cdots\mid U_k,
}
\]
{so \textsc{Reduce} constructs \(U\) and removes the delimiter.}

{In the in-order order, \(U_1\) is completed before \(\textsc{nt}(X)\) is introduced, and \(U_2,\ldots,U_k\) are completed afterward. In the bottom-up order, all \(k\) children are completed before marker introduction. Delimiters opened within children \(U_j\) for \(j>1\) are consumed within those children because their left boundaries differ from that of \(U\). By Lemma~\ref{lem:boundary-lifecycle}, the delimiter at the left boundary of \(U\) is the rightmost active delimiter when \(R_U\) is applied. The reduction domain therefore has the required traversal-specific form. If \(U\) is below its boundary target, \textsc{d-Reduce} constructs \(U\) and retains the delimiter immediately below it; if \(U\) is the boundary target, \textsc{Reduce} constructs \(U\) and consumes the delimiter. The strengthened induction claim follows.}

{The root is the outermost constituent at its left boundary, so its reduction consumes the last retained delimiter. After \(o_{\tau}(T)\), all input has been consumed and the stack contains only \(T\); \textsc{Finish} is therefore applicable. If \(T\) is licensed by \(G\), every construction action instantiates its corresponding production in \(R\), so the same sequence is licensed under \(\Rightarrow_{\tau,G}\).}
\end{proof}

\begin{proposition}[Oracle action count]
\label{prop:oracle-size}
{Let \(T\) contain \(n\) terminals, \(n\) preterminals, and \(m\) phrasal nodes. For every traversal order \(\tau\),}
\[
{
|O_{\tau}(T)|=3n+2m+1.
}
\]
{The top-down oracle introduces exactly \(m\) delimiters, whereas the in-order and bottom-up oracles each introduce}
\[
{
\left|\{\,i:b(w_i)\text{ is defined}\,\}\right|\leq n
}
\]
{delimiters. Every introduced delimiter is consumed by exactly one \textsc{Reduce}.}

{The canonical top-down and in-order oracles have the same action count, whereas the arity-specific canonical bottom-up oracle has \(2n+m+1\) actions.}
\end{proposition}

\begin{proof}
{Each terminal and its preterminal contribute one terminal-consuming action, one \textsc{Nt}(\(P\)), and one \textsc{u-Reduce}, giving \(3n\) actions. Each phrasal node contributes one \textsc{Nt}(\(X\)) and one \textsc{Reduce} or \textsc{d-Reduce}, giving \(2m\) actions. \textsc{Finish} contributes one action. The canonical top-down and in-order systems make the same three terminal-level and two phrase-level decisions. In the canonical bottom-up system, each terminal contributes \textsc{Shift} and \(\textsc{Reduce-}P\#1\), each phrasal node contributes one arity-specific reduction, and \textsc{Finish} contributes one action, giving \(2n+m+1\).}

{Every top-down phrasal \textsc{Nt} introduces a delimiter, giving \(m\) delimiters. In the other two orders, a delimiter is introduced exactly when \(b(w_i)\) is defined. Lemma~\ref{lem:boundary-lifecycle} shows that reductions below the associated boundary target retain it and that reduction of the target consumes it.}
\end{proof}

\begin{definition}[Canonical projection]
\label{def:canonical-projection}
{For \(\tau\in\{\mathrm{TD},\mathrm{IO},\mathrm{BU}\}\), let \(\mathcal{D}_{\tau}\) denote the delimiter-guided system and \(\mathcal{C}_{\tau}\) its canonical counterpart in Appendix~\ref{app:canonical-transition-systems}. In each canonical system, a reduction over a bare terminal is a preterminal construction labeled by \(P\in\mathcal{P}\), a reduction over completed tree items is a phrasal construction labeled by \(X\in\mathcal{N}\), and an unfinished preterminal construction must be completed before any other construction begins. For top-down and in-order action--configuration sequences, \(\Pi_{\tau}\) erases delimiters from every configuration and transforms actions as follows. In the top-down order, it preserves \textsc{Nt}(\(L\)), \textsc{Shift}, and \textsc{Finish}, and maps both \textsc{u-Reduce} and phrase-level \textsc{Reduce} to canonical \textsc{Reduce}. In the in-order order, it maps both \textsc{Shift} and \textsc{d-Shift} to canonical \textsc{Shift}, preserves \textsc{Nt}(\(L\)) and \textsc{Finish}, and maps \textsc{u-Reduce}, \textsc{Reduce}, and \textsc{d-Reduce} to canonical \textsc{Reduce}.}

{In the bottom-up order, \(\Pi_{\mathrm{BU}}\) is a macro projection on successful action--configuration sequences. At completed-construction boundaries it erases delimiters from the stack, and it contracts}
\[
{
(\textsc{Shift}\ \text{or}\ \textsc{d-Shift})
\cdot\textsc{Nt}(P)\cdot\textsc{u-Reduce}
}
\]
{to canonical \(\textsc{Shift}\cdot\textsc{Reduce-}P\#1\), and contracts}
\[
{
\textsc{Nt}(X)\cdot
(\textsc{Reduce}\ \text{or}\ \textsc{d-Reduce})
}
\]
{to \(\textsc{Reduce-}X\#k\), where \(k\) is the number of completed children in the delimited reduction domain. It preserves \textsc{Finish}.}
\end{definition}

\begin{theorem}[Global derivational correspondence]
\label{thm:global-correspondence}
{For every traversal order \(\tau\in\{\mathrm{TD},\mathrm{IO},\mathrm{BU}\}\), \(\Pi_{\tau}\) maps every successful derivation of \(\mathcal{D}_{\tau}\) to a successful derivation of \(\mathcal{C}_{\tau}\) with the same input and final constituency tree. Conversely, every successful derivation \(\delta\) of \(\mathcal{C}_{\tau}\) has a successful delimiter-guided lifting \(\Lambda_{\tau}(\delta)\) that produces the same tree and satisfies}
\[
{
\Pi_{\tau}\bigl(\Lambda_{\tau}(\delta)\bigr)=\delta.
}
\]
{Both directions hold for the structural relations and their grammar-licensed restrictions. Writing \(\mathcal{L}\) and \(\mathcal{L}_{G}\) for the sets of final trees generated under the structural and grammar-licensed relations, respectively,}
\[
{
\mathcal{L}(\mathcal{D}_{\tau})
=
\mathcal{L}(\mathcal{C}_{\tau})
\qquad\text{and}\qquad
\mathcal{L}_{G}(\mathcal{D}_{\tau})
=
\mathcal{L}_{G}(\mathcal{C}_{\tau}).
}
\]
\end{theorem}

\begin{proof}
{Consider first a successful delimiter-guided derivation. In the top-down order, erasing \(D\) from a phrasal source suffix}
\[
{
D\mid\textsc{nt}(X)\mid\alpha_1\mid\cdots\mid\alpha_k
}
\]
{gives exactly the canonical top-down reduction domain. A \textsc{u-Reduce} likewise has the same stack effect as the canonical reduction that constructs its preterminal. All other projected actions have identical marker, terminal, or completion effects. Induction on the derivation therefore gives a successful canonical derivation with the same completed items and final tree.}

{In the in-order order, \textsc{Shift} and \textsc{d-Shift} have the same effect after delimiter erasure. Erasing \(D\) from}
\[
{
D\mid\alpha_1\mid\textsc{nt}(X)\mid
\alpha_2\mid\cdots\mid\alpha_k
}
\]
{gives the canonical in-order reduction domain with the same mother label and ordered children, whether the delimiter-guided reduction retains or removes \(D\). Preterminal reductions project in the same way. Induction again preserves every completed item and the final tree.}

{In a successful bottom-up derivation, the applicability conditions force every consumed bare terminal to be completed by the following \(\textsc{Nt}(P)\cdot\textsc{u-Reduce}\). Once a bottom-up phrasal marker has been introduced, any intervening action would place material above it and make its reduction schema permanently unavailable; hence every successful derivation discharges that marker immediately by \textsc{Reduce} or \textsc{d-Reduce}. The derivation therefore factors into the blocks specified by \(\Pi_{\mathrm{BU}}\). Each projected preterminal block constructs the same \(P(w_i)\), and each projected \(\textsc{Reduce-}X\#k\) consumes the same ordered children and constructs the same mother item. Induction over these blocks yields a successful canonical derivation with the same final tree.}

{For the converse, fix a successful typed canonical derivation and its final tree. Terminals are shifted from left to right, and every construction consumes a complete top suffix. The canonical stack schemas therefore force the nodes of the fixed tree to be constructed in the corresponding recursive order: top-down marker introduction, recursive in-order construction, or recursive postorder construction. Bypassing a required marker or child would bury it below a yield-bearing item outside its reduction domain, after which it could not occupy the required suffix. The following constructions define \(\Lambda_{\tau}\). In the top-down order, retain every canonical preterminal construction, replacing its closing \textsc{Reduce} with \textsc{u-Reduce}. For every phrasal \textsc{Nt}(\(X\)), insert \(D\) immediately before its marker and use phrase-level \textsc{Reduce} to remove that delimiter when the canonical marker is closed. The canonical nesting of markers makes each inserted delimiter the rightmost active delimiter at the matching reduction. Erasing the delimiters and identifying the two delimiter-guided reduction types with canonical \textsc{Reduce} recovers the original derivation.}

{In the in-order order, replace the canonical \textsc{Shift} consuming \(w_i\) with \textsc{d-Shift} exactly when \(w_i\) is the leftmost terminal of at least one phrasal constituent in the final tree; otherwise retain \textsc{Shift}. Preserve each \textsc{Nt}(\(L\)), replace a canonical reduction over a bare terminal with \textsc{u-Reduce}, and replace each phrase-level reduction by \textsc{d-Reduce} for a non-outermost member of its common-left-boundary chain and by \textsc{Reduce} for the outermost member. The canonical schemas impose the same node-construction order as the recursive in-order traversal. Lemma~\ref{lem:boundary-lifecycle} therefore ensures that every selected reduction is applicable and that every introduced delimiter is consumed. Projection action by action recovers the canonical derivation.}

{In a successful typed canonical bottom-up derivation, a \textsc{Shift} introducing \(w_i\) is immediately followed by \(\textsc{Reduce-}P\#1\): if another item were pushed first, the bare terminal could no longer form the top reduction domain required for its preterminal. Replace this pair with \textsc{Shift} or \textsc{d-Shift}, followed by \(\textsc{Nt}(P)\cdot\textsc{u-Reduce}\), using \textsc{d-Shift} exactly when \(w_i\) is the leftmost terminal of at least one phrasal constituent. Replace every phrase-level \(\textsc{Reduce-}X\#k\) with \textsc{Nt}(\(X\)) followed by the \textsc{d-Reduce} or \textsc{Reduce} selected from its common-left-boundary chain. The canonical bottom-up reduction order is the recursive postorder used in Lemma~\ref{lem:boundary-lifecycle}; all lifted blocks are therefore applicable, and no delimiter remains after the root is completed. Each block projects to its original canonical action or action pair, establishing \(\Pi_{\mathrm{BU}}(\Lambda_{\mathrm{BU}}(\delta))=\delta\).}

{Projection and lifting preserve every terminal yield, mother label, and ordered child sequence. They consequently preserve the production instantiated by each construction. The three arguments therefore apply unchanged to the grammar-licensed restrictions, and the two tree-language equalities follow.}
\end{proof}

\section{Experiments}
\label{sec:experiments}
We compare six direct non-binary parsers using phrase-structure annotations from the Wall Street Journal portion of Penn Treebank~3 (PTB) \citep{marcus-etal-1993-building,taylor-marcus-santorini:2003} and from Chinese Treebank~5.1 (CTB) \citep{xue-etal-2005-ctb}. For each traversal order, we train the canonical system and its delimiter-guided counterpart on the same trees with the same neural architecture, optimization protocol, and decoding strategy. The systems differ in how they encode the reduction domain and in the corresponding action inventory, applicability checks, and oracle sequence. The comparison tests whether replacing the canonical domain encoding incurs a loss in labeled constituency accuracy under static-oracle training.

\subsection{Experimental setup}

For each of the three traversal orders---top-down, in-order, and bottom-up---we compare a canonical parser with its delimiter-guided counterpart. All six parsers construct non-binary branching structures directly.
We implement their transitions and static oracles in a shared Stanza training and decoding pipeline \citep{qi-etal-2020-stanza,bauer-manning-2025-high}. All parsers learn from the resulting gold action sequences by teacher forcing. At test time, each parser greedily selects the highest-scoring action permitted by its implementation's structural checks, with no production lookup in the treebank grammar. Appendix~\ref{app:experimental-details} specifies the training and decoding protocol.

Each system--treebank pair has three independent runs (seeds 1111, 2222, and 3333), for a total of 36 runs. We select a checkpoint for each run by labeled development-set \(F_1\). We score its test predictions with \textsc{evalb} \citep{black-etal-1991-procedure} and \textsc{jp-evalb} \citep{jo-etal-2024-novel,park-etal-2024-jp}. Both scorers receive the same saved gold and predicted trees, with matching token yields and reference POS labels in the exported files. The former uses the supplied \texttt{COLLINS.prm} on both corpora. The latter uses its default alignment mode, which retains punctuation. We compute corpus \(F_1\) independently for each seed and report the mean and sample standard deviation of the three scores.

\subsection{Main results}
\label{sec:main-results}

The absolute difference between paired canonical and delimiter-guided means is at most 0.32 \(F_1\) points across the twelve comparisons in Table~\ref{tab:main-results}, with differences computed before rounding. Top-down differences stay within 0.11 points. In-order means are lower with delimiters by 0.14--0.20 points across the two corpora and scorers. For bottom-up parsing, the PTB means differ by less than 0.07 points, while the CTB means rise by 0.31 points under \textsc{evalb} and 0.32 under \textsc{jp-evalb}. The bottom-up pair thus preserves similar labeled constituency accuracy while changing the way the parser identifies a child sequence. These differences describe the observed runs. The experiment specifies no statistical equivalence margin.

\begin{table}[!ht]
\centering
\caption{Labeled test \(F_1\) on PTB and CTB. The published
\textsc{evalb} scores come from a generative top-down parser, a greedy in-order parser trained with a static oracle, and a greedy bottom-up parser trained with a dynamic oracle. These scores reflect the training and evaluation settings of the original studies. 
All six systems in this study use a shared static-oracle training protocol and greedy decoding. Their scores are means with sample standard deviations across three seeds. Dashes indicate unreported
\textsc{jp-evalb} scores.}\label{tab:main-results}
\footnotesize
\resizebox{\textwidth}{!}{
\begin{tabular}{lcccc}
\toprule
& \multicolumn{2}{c}{PTB} & \multicolumn{2}{c}{CTB} \\
\cmidrule(lr){2-3}\cmidrule(lr){4-5}
System & \textsc{evalb} & \textsc{jp-evalb} & \textsc{evalb} & \textsc{jp-evalb} \\
\midrule
\multicolumn{5}{l}{\textit{Previously reported canonical non-binary systems}} \\
\shortstack[l]{Top-down \citep{dyer-etal-2016-recurrent}}
  & 92.40 & --- & 82.70 & --- \\
\shortstack[l]{In-order \citep{liu-zhang-2017-order}}
  & 91.80 & --- & 86.10 & --- \\
\shortstack[l]{Bottom-up \citep{gonzalez-rodriguez-2019-faster}}
  & 91.70 & --- & 86.80 & --- \\
\midrule
\multicolumn{5}{l}{\textit{Canonical non-binary systems, reimplemented}} \\
Top-down & $94.86 \pm 0.07$ & $94.27 \pm 0.09$ & $89.83 \pm 0.28$ & $88.09 \pm 0.31$ \\
In-order & $95.04 \pm 0.07$ & $94.35 \pm 0.08$ & $90.62 \pm 0.19$ & $88.89 \pm 0.20$ \\
Bottom-up & $94.90 \pm 0.04$ & $94.27 \pm 0.09$ & $90.63 \pm 0.27$ & $88.89 \pm 0.26$ \\
\midrule
\multicolumn{5}{l}{\textit{Delimiter-guided systems}} \\
Top-down & $94.89 \pm 0.10$ & $94.25 \pm 0.11$ & $89.90 \pm 0.33$ & $88.20 \pm 0.35$ \\
In-order & $94.84 \pm 0.05$ & $94.18 \pm 0.04$ & $90.48 \pm 0.38$ & $88.74 \pm 0.34$ \\
Bottom-up & $94.90 \pm 0.02$ & $94.21 \pm 0.04$ & $90.94 \pm 0.33$ & $89.20 \pm 0.33$ \\
\bottomrule
\end{tabular}
}
\end{table}

\section{Analysis and discussion}
\label{sec:analysis}

\subsection{Action inventory and derivation workload}
\label{sec:structural-analysis}
\label{sec:action-inventory-analysis}
We count oracle actions on the same gold test trees for each system, including explicit preterminal construction, the \texttt{ROOT} node, and the final \textsc{Finish}. Counting actions on a common output separates derivation length from differences in the structures produced by greedy decoding. Table~\ref{tab:action-inventories} pairs these counts with the action classes observed in the filtered, retagged training data.

\begin{table}[!ht]
\centering
\caption{Action classes observed during training and mean scoring workload per gold test sentence. Actions count oracle decisions. MACs are millions of multiply--accumulate operations in the two affine layers of the action scorer.}
\label{tab:action-inventories}
\small
\setlength{\tabcolsep}{4pt}
\begin{tabular}{lrrrrrr}
\toprule
 & \multicolumn{3}{c}{PTB} & \multicolumn{3}{c}{CTB} \\
\cmidrule(lr){2-4}\cmidrule(lr){5-7}
Order & Classes & Actions & MACs (M) & Classes & Actions & MACs (M) \\
\midrule
\multicolumn{7}{l}{\textit{Canonical}} \\
Top-down & 79 & 110.04 & 17.21 & 65 & 122.15 & 18.67 \\
In-order & 79 & 110.04 & 17.21 & 65 & 122.15 & 18.67 \\
Bottom-up & 251 & 67.25 & 13.48 & 227 & 73.08 & 14.20 \\
\midrule
\multicolumn{7}{l}{\textit{Delimiter-guided}} \\
Top-down & 80 & 110.04 & 17.24 & 66 & 122.15 & 18.70 \\
In-order & 82 & 110.04 & 17.30 & 68 & 122.15 & 18.76 \\
Bottom-up & 82 & 110.04 & 17.30 & 68 & 122.15 & 18.76 \\
\bottomrule
\end{tabular}
\end{table}

Top-down and in-order parsing already locate the reduction domain through a nonterminal marker. Adding delimiters preserves their oracle lengths and adds a small number of classes for preterminal construction and, in in-order parsing, for shifts and reductions with a delimiter effect. The bottom-up change is larger. Canonical bottom-up assigns a separate class to each observed label--arity pair. Delimiter-guided bottom-up uses a label marker and recovers the child sequence from the stack. The vocabulary contracts from 251 to 82 classes on PTB and from 227 to 68 on CTB. Each preterminal and phrasal node now requires a marker action, as in Proposition~\ref{prop:oracle-size}. The resulting gold derivations are 64\% longer on PTB and 67\% longer on CTB.

The narrower output layer and the longer derivation affect different parts of the scoring workload. The decoder computes a score for every class before checking legality. For bottom-up parsing, the number of output scores along a gold derivation falls by 47\% on PTB and 50\% on CTB. The first affine layer, however, maps 532 features to 256 hidden units at every decision in all 36 checkpoints. Counting both affine layers therefore gives the opposite direction. The mean number of multiply--accumulate operations in the bottom-up action scorer rises by 28\% on PTB and 32\% on CTB. The extra decisions repeat the hidden computation as well as the smaller output projection.

The workload calculation uses formal oracle lengths, including for test trees that require an action class absent from the learned inventory. Encoding, composition, stack operations, and legality checks contribute separately to decoding cost. In particular, several delimiter checks scan the current stack, so the number of actions does not determine the cost of those checks.

\subsection{Decoding speed}
\label{sec:decoding-speed}

We benchmark greedy decoding at batch size one on AMD EPYC 7352 cores. For each corpus we sample 20 test sentences without replacement from each of five token-length bands (1--10, 11--20, 21--30, 31--40, and 41 or more). All 36 checkpoints receive the same 100 sentences per corpus, in the same order and with the same predicted POS tags. We load the checkpoints in one Stanza source tree supporting all six systems. Each corpus--seed block runs its six parsers sequentially, pinned to one core with one inference thread. The six blocks run concurrently on cores with separate L3 caches. We rotate the traversal order across seeds and alternate which encoding runs first within a pair.

For each checkpoint we warm up on five additional sentences, one per length band, and time one pass over the 100-sentence sample. The wall-clock interval covers encoding of the pretagged sentence, action scoring, legality checks, and tree construction. POS tagging, checkpoint loading, warm-up, export, and evaluation are outside the timed interval. All decodes completed with the reference token yield. After restoring reference POS tags for comparison, 3,599 of the 3,600 parses matched their archived predictions exactly.

\begin{table}[!ht]
\centering
\caption{Single-core greedy decoding throughput on the length-balanced test samples, in sentences per second. For each checkpoint, throughput is the number of timed sentences divided by their total elapsed parsing time. Entries give the mean and sample standard deviation across the three seeds.}
\label{tab:decoding-throughput}
\small
\begin{tabular}{lcccc}
\toprule
& \multicolumn{2}{c}{PTB} & \multicolumn{2}{c}{CTB} \\
\cmidrule(lr){2-3}\cmidrule(lr){4-5}
Order & Canonical & Delimiter-guided & Canonical & Delimiter-guided \\
\midrule
Top-down & $0.92 \pm 0.04$ & $0.90 \pm 0.03$ & $1.61 \pm 0.01$ & $1.49 \pm 0.04$ \\
In-order & $0.92 \pm 0.10$ & $0.93 \pm 0.09$ & $1.58 \pm 0.05$ & $1.54 \pm 0.06$ \\
Bottom-up & $1.00 \pm 0.06$ & $0.91 \pm 0.06$ & $1.83 \pm 0.05$ & $1.55 \pm 0.03$ \\
\bottomrule
\end{tabular}
\end{table}

Table~\ref{tab:decoding-throughput} reports throughput for the length-balanced samples. On PTB, top-down and in-order throughput changes by at most 0.02 sentences per second when delimiters are introduced. On CTB, the mean falls from 1.61 to 1.49 sentences per second for top-down parsing and from 1.58 to 1.54 for in-order parsing. Canonical bottom-up has the highest mean throughput on both corpora. Delimiter-guided bottom-up reduces throughput by 9\% on PTB and 15\% on CTB relative to its canonical counterpart.

\begin{figure}[!ht]
\centering
\includegraphics[width=\linewidth]{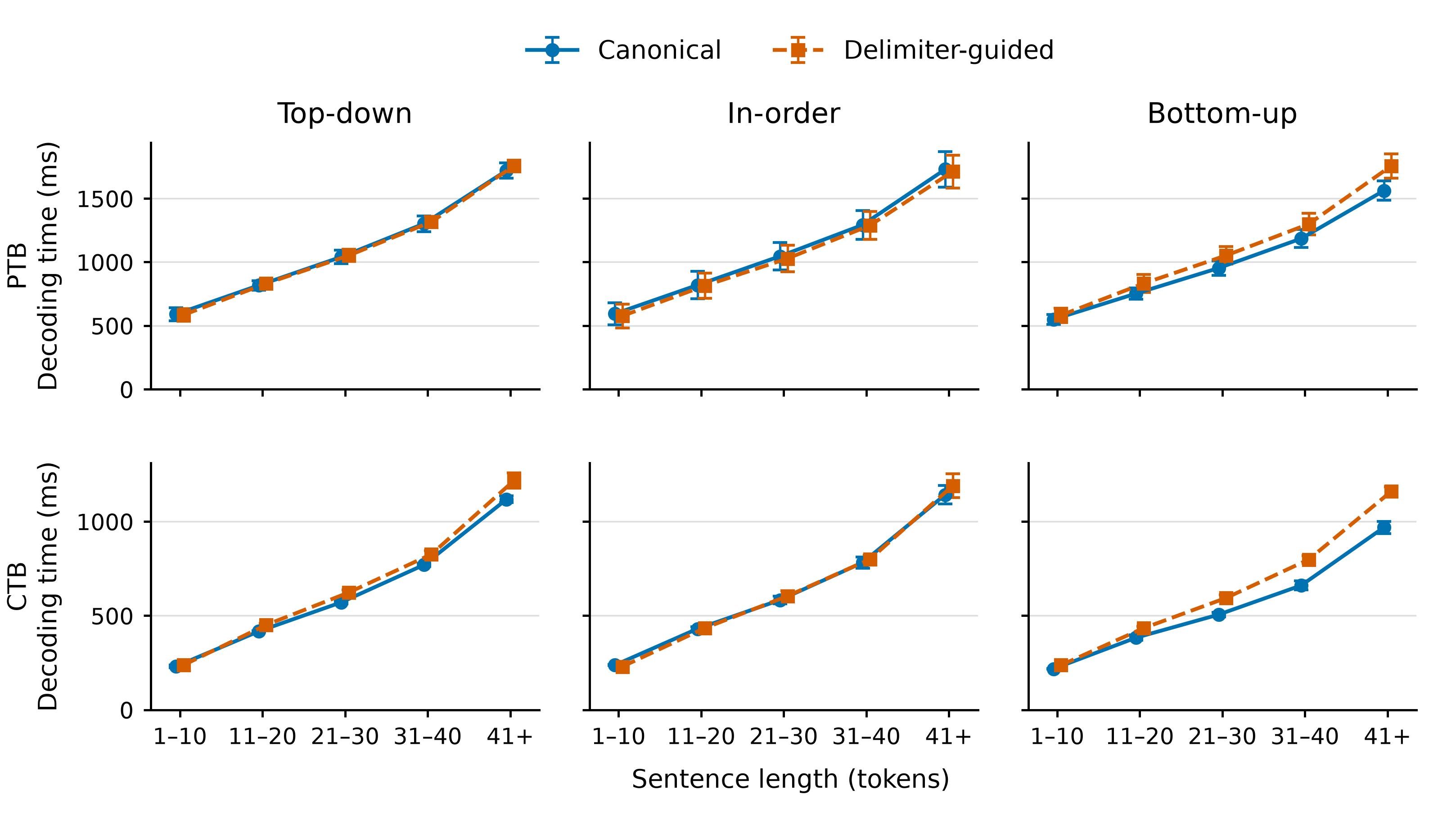}
\caption{Single-core decoding time by sentence length. We average the wall-clock time over the same 20 sentences per band for each checkpoint. Points show the means across seeds and error bars show one sample standard deviation. The two corpus rows have separate vertical scales.}
\label{fig:decoding-latency}
\end{figure}

Mean decoding time increases with sentence length for all six parsers on both corpora (Figure~\ref{fig:decoding-latency}). Delimiter-guided bottom-up takes longer than canonical bottom-up in every length band. For sentences with 41 or more tokens, its mean time rises from 1.56 to 1.76 seconds on PTB and from 0.97 to 1.16 seconds on CTB. These gaps have the same direction as the scorer workload differences in Table~\ref{tab:action-inventories}. The measured time includes encoding and all decoding operations, so it does not isolate the contribution of any one component.

\subsection{Accuracy by constituent arity}
\label{sec:arity-analysis}

We compare the six systems on their recovery of constituents of different arities. For this diagnostic we match phrasal occurrences within each test sentence by label, terminal span, and exact immediate child count. We retain punctuation, omit preterminals and the outer \texttt{ROOT} wrapper, and count repeated occurrences with multiplicity. Exact arity is required even in the group with five or more children. We sum matches and gold and predicted occurrences within each group to compute corpus \(F_1\) per seed.

\begin{table}[!ht]
\centering
\caption{Exact-arity labeled test \(F_1\) by immediate child count. Gold counts give the number of reference phrasal occurrences in each group. Results are means \(\pm\) sample standard deviations across three seeds. C denotes canonical and D delimiter-guided.}
\label{tab:arity-analysis}
\small
\setlength{\tabcolsep}{4pt}
\begin{tabular}{lccccc}
\toprule
System & \(1\) & \(2\) & \(3\) & \(4\) & \(\geq5\) \\
\midrule
\multicolumn{6}{l}{\textit{PTB}} \\
Gold count & 8,879 & 23,840 & 7,381 & 2,317 & 1,859 \\
C Top-down & $93.48 \pm 0.07$ & $92.88 \pm 0.11$ & $88.46 \pm 0.29$ & $81.29 \pm 0.59$ & $81.28 \pm 0.62$ \\
C In-order & $93.65 \pm 0.13$ & $93.10 \pm 0.06$ & $88.57 \pm 0.16$ & $81.42 \pm 0.12$ & $81.64 \pm 0.43$ \\
C Bottom-up & $93.53 \pm 0.04$ & $92.93 \pm 0.12$ & $88.43 \pm 0.21$ & $81.82 \pm 0.54$ & $81.77 \pm 0.13$ \\
D Top-down & $93.49 \pm 0.07$ & $92.86 \pm 0.14$ & $88.50 \pm 0.38$ & $81.11 \pm 0.27$ & $82.05 \pm 0.58$ \\
D In-order & $93.44 \pm 0.11$ & $92.78 \pm 0.06$ & $88.27 \pm 0.15$ & $81.70 \pm 0.21$ & $82.06 \pm 0.68$ \\
D Bottom-up & $93.46 \pm 0.09$ & $92.85 \pm 0.04$ & $88.19 \pm 0.35$ & $81.48 \pm 0.37$ & $82.10 \pm 0.73$ \\
\midrule
\multicolumn{6}{l}{\textit{CTB}} \\
Gold count & 3,482 & 3,698 & 1,121 & 184 & 235 \\
C Top-down & $89.82 \pm 0.18$ & $87.33 \pm 0.23$ & $80.25 \pm 1.30$ & $60.46 \pm 1.53$ & $68.31 \pm 2.28$ \\
C In-order & $90.48 \pm 0.15$ & $88.46 \pm 0.29$ & $81.40 \pm 1.04$ & $69.69 \pm 2.35$ & $74.19 \pm 1.74$ \\
C Bottom-up & $90.53 \pm 0.12$ & $88.42 \pm 0.61$ & $81.80 \pm 0.84$ & $70.77 \pm 1.18$ & $74.44 \pm 0.48$ \\
D Top-down & $89.62 \pm 0.52$ & $87.60 \pm 0.33$ & $80.16 \pm 0.39$ & $62.55 \pm 2.66$ & $69.01 \pm 1.28$ \\
D In-order & $90.18 \pm 0.60$ & $88.34 \pm 0.21$ & $80.63 \pm 0.66$ & $67.47 \pm 1.39$ & $72.68 \pm 1.37$ \\
D Bottom-up & $90.62 \pm 0.28$ & $88.80 \pm 0.61$ & $82.08 \pm 0.76$ & $69.10 \pm 2.33$ & $72.87 \pm 1.58$ \\
\bottomrule
\end{tabular}
\end{table}

All six systems recover unary and binary constituents more accurately than constituents with three or more children. The effect of delimiter guidance varies with traversal order. On PTB, the mean for the group with at least five children rises for each order, by 0.33--0.77 points. On CTB, the four-child mean rises by 2.09 points for top-down parsing and falls by 2.22 for in-order parsing. These two directions hold in all three CTB seed pairs. The bottom-up mean for five or more children falls by 1.57 points. These CTB groups contain 184 and 235 gold occurrences, respectively. The arity results thus identify where the systems differ even when their aggregate scores are close. The criterion jointly measures span, label, and child count.

The canonical bottom-up inventory restricts which label--arity pairs the decoder can predict. Two gold phrasal occurrences on PTB test, an eight-child \texttt{NX} and an eight-child \texttt{PRN}, require label--arity pairs absent from the canonical bottom-up training inventory. No CTB test occurrence has this property. Neither PTB occurrence was recovered by the delimiter-guided bottom-up system in any seed. The development predictions contain examples of successful recovery of unseen pairs. Delimiter-guided bottom-up recovers the 19-child PTB \texttt{NP} and 64-child CTB \texttt{NP} in all three seeds, the 12-child PTB \texttt{VP} in two, and the 15-child CTB \texttt{QP} in none. These examples use the development data also used for checkpoint selection. They illustrate that reusing a known label at a new arity is possible, while accurate recovery still depends on the other construction decisions.

\subsection{Sentence length and agreement between parsers}
\label{sec:length-agreement}

We also group the test sentences by token count and score exact label--span matches, with the same tree treatment as in the arity analysis and with child count omitted from the match. Table~\ref{tab:length-analysis} gives corpus \(F_1\) within each group for both encodings. The shortest group has higher \(F_1\) than the longest group for every system. On PTB, the six means span 95.53--96.05 in the shortest group and 91.99--92.46 in the longest. The corresponding CTB ranges are 93.83--94.88 and 85.66--87.84. For sentences of 41 tokens or more, switching the domain encoding changes the mean by at most 0.29 points on PTB and 0.44 points on CTB within a traversal order. The overall association with sentence length is larger than the observed differences between encodings within this group. Gold derivation workload also varies substantially with length. In the shortest and longest groups, respectively, delimiter-guided bottom-up adds 12.4 and 84.0 oracle actions per sentence on PTB and 11.1 and 118.8 on CTB.

\begin{table}[!ht]
\centering
\caption{Labeled bracket \(F_1\) by test sentence length in tokens. Counts give the number of test sentences in each group. Results are means \(\pm\) sample standard deviations across three seeds. C denotes canonical and D delimiter-guided.}
\label{tab:length-analysis}
\small
\setlength{\tabcolsep}{4pt}
\begin{tabular}{lccccc}
\toprule
System & 1--10 & 11--20 & 21--30 & 31--40 & \(\geq41\) \\
\midrule
\multicolumn{6}{l}{\textit{PTB}} \\
Sentences & 270 & 764 & 778 & 433 & 171 \\
C Top-down & $95.65 \pm 0.19$ & $95.21 \pm 0.20$ & $94.32 \pm 0.12$ & $93.39 \pm 0.14$ & $92.41 \pm 0.43$ \\
C In-order & $95.99 \pm 0.29$ & $95.17 \pm 0.09$ & $94.49 \pm 0.12$ & $93.53 \pm 0.14$ & $92.28 \pm 0.12$ \\
C Bottom-up & $95.53 \pm 0.47$ & $95.06 \pm 0.12$ & $94.60 \pm 0.06$ & $93.19 \pm 0.16$ & $92.34 \pm 0.39$ \\
D Top-down & $96.05 \pm 0.09$ & $95.16 \pm 0.07$ & $94.21 \pm 0.15$ & $93.43 \pm 0.23$ & $92.46 \pm 0.06$ \\
D In-order & $95.94 \pm 0.48$ & $95.10 \pm 0.08$ & $94.33 \pm 0.15$ & $93.28 \pm 0.14$ & $91.99 \pm 0.09$ \\
D Bottom-up & $95.67 \pm 0.23$ & $95.07 \pm 0.09$ & $94.36 \pm 0.13$ & $93.28 \pm 0.08$ & $92.24 \pm 0.11$ \\
\midrule
\multicolumn{6}{l}{\textit{CTB}} \\
Sentences & 96 & 81 & 84 & 38 & 49 \\
C Top-down & $94.11 \pm 0.74$ & $91.64 \pm 0.25$ & $91.50 \pm 0.79$ & $89.60 \pm 0.98$ & $85.99 \pm 1.32$ \\
C In-order & $94.88 \pm 0.87$ & $91.98 \pm 0.70$ & $92.21 \pm 0.50$ & $90.47 \pm 0.72$ & $87.12 \pm 0.13$ \\
C Bottom-up & $93.83 \pm 0.65$ & $92.57 \pm 0.58$ & $91.81 \pm 0.39$ & $90.34 \pm 0.47$ & $87.40 \pm 0.48$ \\
D Top-down & $94.27 \pm 0.40$ & $92.10 \pm 0.20$ & $91.84 \pm 0.59$ & $89.71 \pm 0.48$ & $85.66 \pm 0.50$ \\
D In-order & $94.25 \pm 0.44$ & $92.26 \pm 0.55$ & $91.83 \pm 0.82$ & $90.72 \pm 1.16$ & $86.82 \pm 0.28$ \\
D Bottom-up & $94.71 \pm 0.61$ & $92.61 \pm 0.50$ & $91.99 \pm 0.34$ & $90.77 \pm 0.12$ & $87.84 \pm 1.06$ \\
\bottomrule
\end{tabular}
\end{table}

Close corpus scores allow substantial disagreement on individual sentences. We pair runs with the same seed within each traversal order and test whether the predicted multisets of labeled brackets coincide. Agreement ranges from 60.00\% to 64.67\% across orders on PTB and from 50.86\% to 53.16\% on CTB (Table~\ref{tab:bracket-agreement}). Under sentence-level \(F_1\), both encodings outperform their counterpart on some sentences. For example, on CTB the delimiter-guided bottom-up predictions score higher on an average of 24.04\% of sentences, and the canonical predictions score higher on 23.85\%.

\begin{table}[!ht]
\centering
\caption{Sentence-level agreement and relative accuracy for matched seed pairs (\%, mean \(\pm\) sample standard deviation). Same requires identical labeled bracket multisets. D higher and C higher count sentences with higher bracket \(F_1\) for the delimiter-guided and canonical predictions, respectively.}
\label{tab:bracket-agreement}
\small
\setlength{\tabcolsep}{4pt}
\begin{tabular}{lcccccc}
\toprule
 & \multicolumn{3}{c}{PTB} & \multicolumn{3}{c}{CTB} \\
\cmidrule(lr){2-4}\cmidrule(lr){5-7}
Order & Same & D higher & C higher & Same & D higher & C higher \\
\midrule
Top-down & $64.67 \pm 2.06$ & $16.90 \pm 1.52$ & $16.63 \pm 0.93$ & $52.78 \pm 1.30$ & $22.70 \pm 1.00$ & $23.08 \pm 0.88$ \\
In-order & $62.16 \pm 1.27$ & $17.40 \pm 0.87$ & $18.50 \pm 1.51$ & $53.16 \pm 2.87$ & $20.31 \pm 2.88$ & $24.81 \pm 3.02$ \\
Bottom-up & $60.00 \pm 1.53$ & $18.65 \pm 0.56$ & $19.23 \pm 1.02$ & $50.86 \pm 1.99$ & $24.04 \pm 2.91$ & $23.85 \pm 3.89$ \\
\bottomrule
\end{tabular}
\end{table}

We further count gold bracket occurrences recovered when matching only spans and when requiring labels as well. Across the six systems, 5.35--5.56\% of PTB gold occurrences and 8.34--9.76\% of CTB gold occurrences lack a matching predicted span after matching with multiplicity. Requiring the label removes a further 0.80--0.86\% of gold occurrences on PTB and 1.36--1.59\% on CTB. Missed spans therefore account for the larger share of the observed recall deficit in both encodings. These measurements describe completed trees. They leave the transition at which a parser first diverged from its gold derivation unresolved.

\subsection{Tree-language equivalence and learned decisions}
\label{sec:representational-scope}
\label{sec:neural-structural-control}

Theorem~\ref{thm:global-correspondence} equates the final tree languages of the formal systems by projecting and lifting successful derivations. The lifting uses the completed tree to choose where to place delimiters. For example, the canonical bottom-up parser can shift the second word of a sentence and leave open whether a later constituent begins there. In the formal delimiter-guided system, a plain \textsc{Shift} at that word rules out a phrase beginning there, because no later action can introduce a delimiter at a boundary the parser has already passed. In-order parsing makes the same early commitment, while top-down parsing introduces the boundary jointly with the mother label. A misplaced delimiter can affect several constituents with the same left edge if successive reductions retain it. The bidirectional encoder supplies sentence context in every order, while the traversal order determines which children are complete when the mother label is chosen. Static supervision trains these decisions on gold configurations. 

Across all six systems and three seeds, the audit found zero structural departures in 43,488 PTB parses and 6,264 CTB parses after accounting for top-down marker timing. The relaxed in-order domain and the bottom-up fallback each occurred zero times at marker introduction and reduction. For each of those systems this covers 7,248 PTB parses and 1,044 CTB parses. Delimiter-guided top-down parsing delayed delimiter placement at every phrasal opening. The audit places each delimiter at its corresponding opening action before checking the structural schemas (Appendix~\ref{app:experimental-details}).

Binarization also changes the derivation while allowing the original branching structure to be recovered. It imposes a sequence of local compositions and introduces intermediate training targets \citep{sagae-lavie-2005-classifier,watanabe-sumita-2015-transition,liu-zhang-2017-shift,qi-etal-2026-better}. The direct systems compose the entire annotated child sequence at each phrase-level reduction. The earlier binary bottom-up experiments of \citet{park-etal-2026-non} used a different model configuration and training schedule. Estimating an effect of binarization against the six present systems would require a binary run with the same protocol.

\section{Conclusion}

{This paper identifies the reduction domain as the common structural object of direct non-binary top-down, in-order, and bottom-up parsing. In an applicable phrase-level reduction, the rightmost delimiter identifies the relevant stack segment, the phrasal marker supplies the mother label, and the completed items in that segment supply the ordered children. The reduction arity is therefore the cardinality of an already identified child sequence rather than an independent parameter of the reduction action.}

{The formal results separate this representational claim from grammar licensing. A stack invariant is preserved throughout every structural derivation and supports soundness and completeness under both the structural and grammar-licensed relations. The recursively specified static oracle has an exact action count of \(3n+2m+1\). More centrally, the global projection-and-lifting theorem ranges over complete successful derivations of all three traversal orders. It shows that each delimiter-guided system and its canonical counterpart construct the same final trees, and hence generate the same tree language, under both relations. Delimiter guidance and canonical domain encoding thus change the derivation representation without changing its tree-building capacity.}

{The paired experiments show that the factorization can support accurate greedy parsing under static-oracle training. In bottom-up parsing, deriving arity from the stack reduces the learned action inventory by more than two thirds on both treebanks and increases the number of decisions in a gold derivation.}

\section*{Declaration of Competing Interest}
The authors declare no competing interests.

\section*{Code availability}
The Stanza implementations of all six parsers, along with the notebooks, experiment scripts, and results, are available at \url{https://github.com/jungyeul/generalized-non-binary}.

\section*{Declaration of generative AI and AI-assisted technologies in the manuscript preparation process}

The authors used OpenAI ChatGPT and Codex to edit the language, reorganize parts of the manuscript, and clarify the technical and mathematical exposition. The authors reviewed and edited the resulting text, verified the formal arguments and implementations, and take full responsibility for the article.

\section*{Acknowledgments}
This work was supported by the Institute for Information \& Communications Technology Planning \& Evaluation (RS-2025-25461932, \textit{the Top-Tier AI Global HRD invitation program}, and RS-2022-II220078, \textit{Explainable Logical Reasoning for Medical Knowledge Generation}).


\appendix

\section{Canonical transition systems}
\label{app:canonical-transition-systems}

{For each traversal order \(\tau\in\{\mathrm{TD},\mathrm{IO},\mathrm{BU}\}\), the canonical system \(\mathcal{C}_{\tau}\) has configurations \(c=\langle\sigma,i,f\rangle\), where \(\sigma\) is a stack, \(i\in\{1,\ldots,n+1\}\) is the next-input index, and \(f\in\{0,1\}\) is the completion flag. The initial configuration is \(\langle\epsilon,1,0\rangle\). A successful final configuration has the form \(\langle T,n+1,1\rangle\), where \(T\) is a completed tree rooted in \(S\) with yield \(w_1\cdots w_n\). Top-down and in-order stacks may contain markers \(\textsc{nt}(L)\), bare terminals, and completed tree items; bottom-up stacks may contain bare terminals and completed tree items.}

{Each action below is a partial function on configurations, and their union defines the canonical structural relation \(\Rightarrow^{\mathcal{C}}_{\tau}\). In every schema, \(\sigma\) denotes an arbitrary stack prefix and the displayed elements constitute its complete suffix. The variables \(P\in\mathcal{P}\) and \(X\in\mathcal{N}\) distinguish preterminal and phrasal instances of the unified label variable \(L\in\mathcal{P}\cup\mathcal{N}\). A bare terminal can be reduced only to a preterminal, whereas a nonempty sequence of completed tree items can be reduced only to a phrasal constituent. Every action requires \(f=0\), and no action is applicable after \(f=1\).}

{An unfinished preterminal construction must be completed before any other transition is taken. In the top-down system, a stack-final \(\textsc{nt}(P)\) admits only \textsc{Shift}, and the resulting suffix \(\textsc{nt}(P)\mid w_{i-1}\) admits only \textsc{Reduce}. In the in-order system, a stack-final bare terminal \(w_{i-1}\) admits only \(\textsc{Nt}(P)\), and the resulting suffix \(w_{i-1}\mid\textsc{nt}(P)\) admits only \textsc{Reduce}. In the bottom-up system, a stack-final bare terminal \(w_{i-1}\) admits only \(\textsc{Reduce-}P\#1\).}

\paragraph{Top-down}

{The canonical top-down action inventory consists of \(\textsc{Nt}(L)\), \textsc{Shift}, \textsc{Reduce}, and \textsc{Finish}. The expressions \(\textsc{Nt}(P)\) and \(\textsc{Nt}(X)\) below are typed instances of the single action family \(\textsc{Nt}(L)\), and both reduction schemas are instances of the single action \textsc{Reduce}:}
\[
{
\begin{array}{llll}
\textsc{Nt}(P):&
\langle\sigma,i,0\rangle
&\Rightarrow^{\mathcal{C}}_{\mathrm{TD}}&
\langle\sigma\mid\textsc{nt}(P),i,0\rangle,
\qquad i\leq n,
\\[1mm]
\textsc{Nt}(X):&
\langle\sigma,i,0\rangle
&\Rightarrow^{\mathcal{C}}_{\mathrm{TD}}&
\langle\sigma\mid\textsc{nt}(X),i,0\rangle,
\qquad i\leq n,
\\[1mm]
\textsc{Shift}:&
\langle\sigma\mid\textsc{nt}(P),i,0\rangle
&\Rightarrow^{\mathcal{C}}_{\mathrm{TD}}&
\langle\sigma\mid\textsc{nt}(P)\mid w_i,i+1,0\rangle,
\qquad i\leq n,
\\[1mm]
\textsc{Reduce}:&
\langle\sigma\mid\textsc{nt}(P)\mid w_{i-1},i,0\rangle
&\Rightarrow^{\mathcal{C}}_{\mathrm{TD}}&
\langle\sigma\mid P(w_{i-1}),i,0\rangle,
\qquad 2\leq i\leq n+1,
\\[1mm]
\textsc{Reduce}:&
\langle\sigma\mid\textsc{nt}(X)\mid\alpha_1\mid\cdots\mid\alpha_k,i,0\rangle
&\Rightarrow^{\mathcal{C}}_{\mathrm{TD}}&
\langle\sigma\mid X(\alpha_1,\ldots,\alpha_k),i,0\rangle,
\qquad k\geq1.
\end{array}
}
\]
{In a phrase-level \textsc{Reduce}, the displayed \(\textsc{nt}(X)\) is the topmost unmatched marker, and \(\alpha_1,\ldots,\alpha_k\) are completed tree items with nonempty consecutive yields in left-to-right order. The marker supplies the mother label, and the complete suffix above it supplies the ordered child sequence.}

\paragraph{In-order}

{The canonical in-order action inventory likewise consists of \textsc{Shift}, \(\textsc{Nt}(L)\), \textsc{Reduce}, and \textsc{Finish}. Its typed transition schemas are}
\[
{
\begin{array}{llll}
\textsc{Shift}:&
\langle\sigma,i,0\rangle
&\Rightarrow^{\mathcal{C}}_{\mathrm{IO}}&
\langle\sigma\mid w_i,i+1,0\rangle,
\qquad i\leq n,
\\[1mm]
\textsc{Nt}(P):&
\langle\sigma\mid w_{i-1},i,0\rangle
&\Rightarrow^{\mathcal{C}}_{\mathrm{IO}}&
\langle\sigma\mid w_{i-1}\mid\textsc{nt}(P),i,0\rangle,
\qquad 2\leq i\leq n+1,
\\[1mm]
\textsc{Reduce}:&
\langle\sigma\mid w_{i-1}\mid\textsc{nt}(P),i,0\rangle
&\Rightarrow^{\mathcal{C}}_{\mathrm{IO}}&
\langle\sigma\mid P(w_{i-1}),i,0\rangle,
\qquad 2\leq i\leq n+1,
\\[1mm]
\textsc{Nt}(X):&
\langle\sigma\mid\alpha_1,i,0\rangle
&\Rightarrow^{\mathcal{C}}_{\mathrm{IO}}&
\langle\sigma\mid\alpha_1\mid\textsc{nt}(X),i,0\rangle,
\\[1mm]
\textsc{Reduce}:&
\langle\sigma\mid\alpha_1\mid\textsc{nt}(X)\mid\alpha_2\mid\cdots\mid\alpha_k,i,0\rangle
&\Rightarrow^{\mathcal{C}}_{\mathrm{IO}}&
\langle\sigma\mid X(\alpha_1,\ldots,\alpha_k),i,0\rangle,
\qquad k\geq1.
\end{array}
}
\]
{In a phrase-level \textsc{Reduce}, the displayed marker is the topmost unmatched marker, and \(\alpha_1,\ldots,\alpha_k\) are completed tree items with nonempty consecutive yields in left-to-right order. The sequence \(\alpha_2,\ldots,\alpha_k\) is empty when \(k=1\), so unary phrasal constituents remain distinct from preterminal constructions by the type of their sole child.}

\paragraph{Arity-specific bottom-up}

{The canonical arity-specific bottom-up action inventory consists of \textsc{Shift}, \(\textsc{Reduce-}L\#k\), and \textsc{Finish}. The preterminal and phrasal schemas are typed instances of the single arity-specific reduction family:}
\[
{
\begin{array}{llll}
\textsc{Shift}:&
\langle\sigma,i,0\rangle
&\Rightarrow^{\mathcal{C}}_{\mathrm{BU}}&
\langle\sigma\mid w_i,i+1,0\rangle,
\qquad i\leq n,
\\[1mm]
\textsc{Reduce-}P\#1:&
\langle\sigma\mid w_{i-1},i,0\rangle
&\Rightarrow^{\mathcal{C}}_{\mathrm{BU}}&
\langle\sigma\mid P(w_{i-1}),i,0\rangle,
\qquad 2\leq i\leq n+1,
\\[1mm]
\textsc{Reduce-}X\#k:&
\langle\sigma\mid\alpha_1\mid\cdots\mid\alpha_k,i,0\rangle
&\Rightarrow^{\mathcal{C}}_{\mathrm{BU}}&
\langle\sigma\mid X(\alpha_1,\ldots,\alpha_k),i,0\rangle,
\qquad k\geq1.
\end{array}
}
\]
{In the phrase-level schema, \(\alpha_1,\ldots,\alpha_k\) are completed tree items with nonempty consecutive yields in left-to-right order. The parameter \(k\) selects the complete top suffix and records its arity. In particular, \(\textsc{Reduce-}P\#1\) applies only to a bare terminal, whereas \(\textsc{Reduce-}X\#1\) applies only to a completed tree item.}

\paragraph{Completion and grammar licensing}

{For every traversal order, \textsc{Finish} applies exactly when all input has been consumed and the stack contains one completed tree rooted in \(S\):}
\[
{
\begin{array}{llll}
\textsc{Finish}:&
\langle T,n+1,0\rangle
&\Rightarrow^{\mathcal{C}}_{\tau}&
\langle T,n+1,1\rangle,
\qquad \operatorname{root}(T)=S.
\end{array}
}
\]

{The canonical grammar-licensed relation \(\Rightarrow^{\mathcal{C}}_{\tau,G}\subseteq\Rightarrow^{\mathcal{C}}_{\tau}\) requires every constructed constituent to instantiate a production in \(R\). A preterminal reduction constructing \(P(w_{i-1})\) requires \(P\rightarrow w_{i-1}\in R\). A phrase-level reduction constructing \(X(\alpha_1,\ldots,\alpha_k)\) requires \(X\rightarrow Y_1\cdots Y_k\in R\), where \(Y_j\) is the root label of \(\alpha_j\). All nonconstructive transitions coincide under the structural and grammar-licensed relations.}

\section{Illustrative oracle derivations}
\label{sec:oracle-comparison}

\begin{figure}[!ht]
\centering\footnotesize
\begin{forest}
for tree={
align=center,
rounded corners,
if n children=0{
tier=word
}{}
}
[TOP
[S
[NP$_1$
[NNP [Ms.]]
[NNP [Haag]]
]
[VP
[VBZ [plays]]
[NP$_2$
[NNP [Elianti]]
]
]
[. [.]]
]
]
\end{forest}
\caption{Target constituency tree for
\textit{Ms.~Haag plays Elianti.}}
\label{fig:ms-haag}
\end{figure}
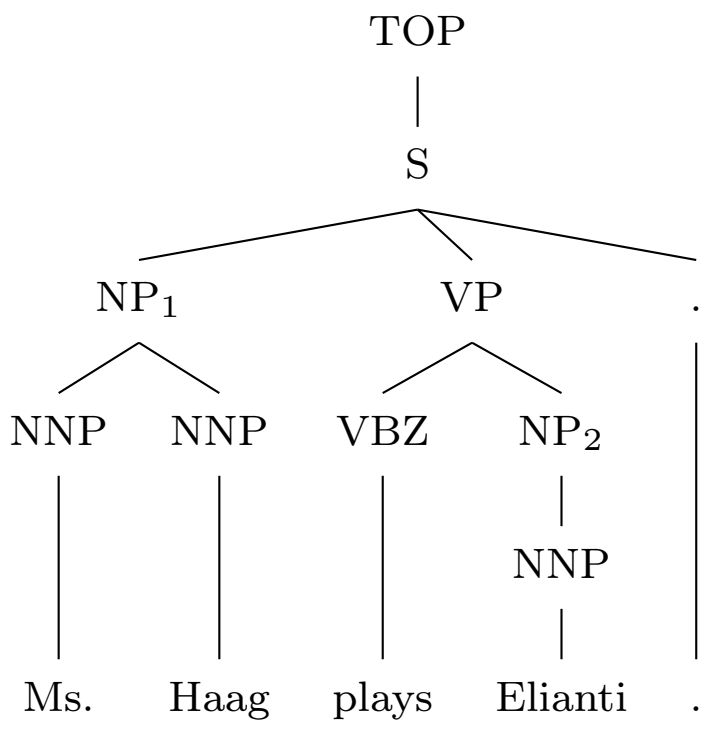

Figure~\ref{fig:ms-haag} gives the target tree for
\textit{Ms.~Haag plays Elianti.}, with \(TOP\) treated as a phrasal
nonterminal. The tree has five terminals, five preterminals, and five
phrasal nodes, with phrase-level arities one, two, and three.

Table~\ref{tab:three-oracles} gives the static oracle sequence for each
traversal order. Every sequence contains 26 actions, including
\textsc{Finish}, and is a successful derivation of the tree in
Figure~\ref{fig:ms-haag} under
\(\Rightarrow_{\tau}\). If the displayed tree is licensed
by \(G\), the same sequences are derivations under
\(\Rightarrow_{\tau,G}\). In each phrase-level construction, the marker
follows exactly \(\operatorname{Order}_{\tau}(k)\) completed children.
{In the top-down column, each phrasal \(\textsc{Nt}(X)\) is a single action whose stack effect introduces \(D\mid\textsc{nt}(X)\); hence a delimiter is present even though no \textsc{d-Shift} action appears. Preterminal \(\textsc{Nt}(P)\) introduces only \(\textsc{nt}(P)\) and no delimiter.}

\begin{table}[!ht]
\centering
\caption{Delimiter-guided static-oracle sequences for the tree in Figure~\ref{fig:ms-haag}. Tokens shown with \textsc{Shift} and
\textsc{d-Shift} identify the consumed input token; they are not action
parameters. The reduction actions have neither label nor arity
parameters.}
\footnotesize
\label{tab:three-oracles}
\begin{tabular}{rlll}
\toprule
\textbf{Step} & \textbf{Top-down} & \textbf{In-order} & \textbf{Bottom-up} \\
\midrule
1& \textsc{Nt}(TOP) & \textsc{d-Shift}(Ms.) & \textsc{d-Shift}(Ms.) \\
2& \textsc{Nt}(S) & \textsc{Nt}(NNP)& \textsc{Nt}(NNP) \\
3& \textsc{Nt}(NP$_1$)& \textsc{u-Reduce} & \textsc{u-Reduce} \\
4& \textsc{Nt}(NNP) & \textsc{Nt}(NP$_1$) & \textsc{Shift}(Haag) \\
5& \textsc{Shift}(Ms.)& \textsc{Shift}(Haag)& \textsc{Nt}(NNP) \\
6& \textsc{u-Reduce}& \textsc{Nt}(NNP)& \textsc{u-Reduce} \\
7& \textsc{Nt}(NNP) & \textsc{u-Reduce} & \textsc{Nt}(NP$_1$) \\
8& \textsc{Shift}(Haag) & \textsc{d-Reduce} & \textsc{d-Reduce} \\
9& \textsc{u-Reduce}& \textsc{Nt}(S)& \textsc{d-Shift}(plays) \\
10 & \textsc{Reduce}& \textsc{d-Shift}(plays) & \textsc{Nt}(VBZ) \\
11 & \textsc{Nt}(VP)& \textsc{Nt}(VBZ)& \textsc{u-Reduce} \\
12 & \textsc{Nt}(VBZ) & \textsc{u-Reduce} & \textsc{d-Shift}(Elianti) \\
13 & \textsc{Shift}(plays)& \textsc{Nt}(VP) & \textsc{Nt}(NNP) \\
14 & \textsc{u-Reduce}& \textsc{d-Shift}(Elianti) & \textsc{u-Reduce} \\
15 & \textsc{Nt}(NP$_2$)& \textsc{Nt}(NNP)& \textsc{Nt}(NP$_2$) \\
16 & \textsc{Nt}(NNP) & \textsc{u-Reduce} & \textsc{Reduce} \\
17 & \textsc{Shift}(Elianti)& \textsc{Nt}(NP$_2$) & \textsc{Nt}(VP) \\
18 & \textsc{u-Reduce}& \textsc{Reduce}& \textsc{Reduce} \\
19 & \textsc{Reduce}& \textsc{Reduce}& \textsc{Shift}(.) \\
20 & \textsc{Reduce}& \textsc{Shift}(.)& \textsc{Nt}(.) \\
21 & \textsc{Nt}(.) & \textsc{Nt}(.) & \textsc{u-Reduce} \\
22 & \textsc{Shift}(.)& \textsc{u-Reduce}& \textsc{Nt}(S) \\
23 & \textsc{u-Reduce}& \textsc{d-Reduce}& \textsc{d-Reduce} \\
24 & \textsc{Reduce}& \textsc{Nt}(TOP) & \textsc{Nt}(TOP) \\
25 & \textsc{Reduce}& \textsc{Reduce}& \textsc{Reduce} \\
26 & \textsc{Finish}& \textsc{Finish}& \textsc{Finish} \\
\bottomrule
\end{tabular}
\end{table}

For the in-order and bottom-up sequences, the boundary targets are
\[
b(\textit{Ms.})=TOP,\qquad
b(\textit{plays})=VP,\qquad
b(\textit{Elianti})=NP_2,
\]
while \(b(\textit{Haag})\) and the boundary target of the final
punctuation are undefined. Hence, both sequences introduce delimiters
before \textit{Ms.}, \textit{plays}, and \textit{Elianti}. The delimiter
before \textit{Ms.} is preserved when \(NP_1\) and \(S\) are constructed
and consumed when its boundary target \(TOP\) is constructed. The other
two delimiters are consumed when \(VP\) and \(NP_2\), respectively, are
constructed.

Immediately before constructing the ternary \(S\), the three reducible
suffixes are
\[
\begin{array}{ll}
\mathrm{TD}:&
D\mid\textsc{nt}(S)\mid NP_1\mid VP\mid {.}
\\[1mm]
\mathrm{IO}:&
D\mid NP_1\mid\textsc{nt}(S)\mid VP\mid {.}
\\[1mm]
\mathrm{BU}:&
D\mid NP_1\mid VP\mid {.}\mid\textsc{nt}(S)
\end{array}
\]
Deleting the delimiter and marker yields the same ordered reduction
domain
\[
\langle NP_1,VP,{.}\rangle
\]
in all three cases, and hence the same recovered arity \(k=3\). The
top-down sequence constructs \(S\) with \textsc{Reduce}; the in-order
and bottom-up sequences use \textsc{d-Reduce} because the active
delimiter must remain available for the subsequent construction of
\(TOP\).

\section{Experimental details}
\label{app:experimental-details}

\paragraph{Data and preprocessing}
The PTB split assigns WSJ sections 02--21 to training, section 22 to development, and section 23 to testing. The prepared CTB 5.1 splits follow the basic partition, which assigns documents 001--270 and 440--1151 to training, 301--325 to development, and 271--300 to testing. The runners consume prepared split files. They prune null elements, discard any tree made empty by pruning, simplify functional and coindexation labels, and standardize the outer label to \texttt{ROOT} with one child. The resulting train, development, and test files contain 39,832, 1,700, and 2,416 trees for PTB and 17,544, 352, and 348 for CTB. The trainer removes duplicate trees from the training and development files, giving 39,514 and 1,696 trees for PTB and 16,676 and 326 for CTB. The prepared files for each split have identical content hashes across systems and seeds. The test runs score all 2,416 PTB trees and all 348 CTB trees.

\paragraph{Encoder and action scorer}
The experiments use a modified Stanza 1.14.0 parser with Python 3.10.12, PyTorch \texttt{2.14.1+cu130} and Transformers 4.57.6. The word encoder concatenates pretrained word vectors, trainable word and POS embeddings, forward and backward character language-model features, and a learned linear mixture of the final four hidden layers of frozen BERT \citep{devlin-etal-2019-bert}. PTB uses \texttt{bert-large-cased}, Stanza's \texttt{conll17} vectors, and the \texttt{1billion} character models. CTB uses \texttt{bert-base-chinese}, \texttt{fasttext157} vectors, and the \texttt{gigaword} character models. A two-layer bidirectional LSTM encodes the words with 256 units per direction. The constituent and transition stacks use two-layer LSTMs with hidden sizes 256 and 20, respectively. Transition and POS embeddings have 20 dimensions, and the trainable word embeddings have 100. Constituent composition takes a coordinatewise maximum over child vectors, followed by a learned transformation. Two affine layers with a ReLU hidden layer of width 256 score the actions. Word, prediction, and LSTM input dropout are 0.2. The models use no additional attention layers.

\paragraph{Training and model selection}
Before oracle construction, Stanza's default POS pipelines assign XPOS tags to the supplied tokens (\texttt{combined\_charlm} for English and \texttt{gsdsimp\_charlm} for Chinese). These tags serve as encoder features and as preterminal labels in the training trees. The transition sequences shift bare words and predict the preterminal label through the respective construction actions. Training minimizes the sum of action cross-entropies over each minibatch, with the softmax taken over the learned action inventory. Each of the 200 training iterations takes the next 5,000 trees from shuffled passes through the training set, carrying unused trees forward between iterations. Training batches contain 16 trees and evaluation batches contain 32. We use AdamW with initial learning rate 0.001, \(\beta_1=0.9\), \(\beta_2=0.999\), and weight decay 0.05 (zero for LSTM biases) in a single training stage. A development-score plateau scheduler multiplies the learning rate by 0.6 after a patience of five iterations, with a ten-iteration cooldown and a floor of \(2\times10^{-5}\). We retain the earliest checkpoint attaining the highest native labeled development \(F_1\), computed with CoreNLP 4.6.0 on the filtered, retagged development trees. The canonical top-down PTB runs resumed from optimizer and scheduler checkpoints after 114, 132, and 132 iterations for seeds 1111, 2222, and 3333, respectively. The sampler and random-number generators were reinitialized on resume. All three runs continued through iteration 200.

\paragraph{Implementation checks at decoding}
The learned policy scores the observed training action classes and uses order-specific legality checks, including root restrictions and safeguards against unbounded action sequences. The unary limits derived from training are 4 for PTB and 6 for CTB. Both top-down implementations resolve a pending marker's role from the next executed action. A shift identifies a preterminal marker, and a child marker identifies a phrasal marker. In the delimiter-guided implementation, the latter action also materializes the parent's delimiter. The delimiter-guided in-order checks allow a phrasal marker with multiple completed items to its left in the active segment. The delimiter-guided bottom-up checks allow the entire completed stack as the reduction domain when no delimiter is active. All reported test predictions cover the full reference token sequences.

When no delimiter is active, the in-order decoder allows \textsc{d-Shift} only from an empty stack. The bottom-up decoder disallows \textsc{d-Shift} while a marker is at the top of the stack. In both orders, \textsc{d-Reduce} requires an active delimiter and a nonroot label. A unary \textsc{d-Reduce} is subject to the decoder's unary guard. When the delimiter is at the bottom of the in-order stack, only a root reduction may consume it.

\paragraph{Audit of executed transitions}
We rerun greedy decoding for the 36 test checkpoints on an NVIDIA A100 with length-sorted inputs and the original evaluation batch size of 32. We record the selected actions and replay them against the structural schemas in Section~\ref{sec:generalized-transition-operators} and Appendix~\ref{app:canonical-transition-systems}. For top-down traces, we assign each marker its resolved type at its opening action and, in the delimiter-guided variant, place the delimiter at the same action. This moves 139,023 PTB delimiters and 26,917 CTB delimiters one action earlier in the recorded histories. For delimiter-guided in-order and bottom-up traces, we check for the respective relaxed domain at both marker introduction and reduction. The symbolic replay reconstructs every decoded tree from a total of 5,177,168 actions. All 49,752 decoded trees match the archived predictions after reference POS tags are restored.

\paragraph{Test scoring}
We restore the reference POS labels after decoding and retain the saved \texttt{ROOT} wrappers for both scorers. We run \textsc{evalb} with the \texttt{COLLINS.prm} file distributed with \textsc{jp-evalb}, applying
the same label exclusions and equivalences to PTB and CTB. We run \textsc{jp-evalb} in its default alignment mode, which retains punctuation. The scoring records identify the exact revisions of both scorers. For each seed, we compute corpus-level \(F_1\) from the matched, predicted, and gold bracket totals. We round the mean and sample standard deviation only after aggregating across seeds.

\paragraph{Data provenance}
The Linguistic Data Consortium distributes both corpora under license.
We use Penn Treebank~3 (\texttt{LDC99T42}) and the Chinese Treebank~5.1
update to \texttt{LDC2005T01}.

\doublespacing

\end{document}